\pdfoutput=1

\documentclass[11pt]{article}

\usepackage{acl}

\usepackage{amssymb}
\usepackage{pifont}

\usepackage{times}
\usepackage{latexsym}
\usepackage{tikz, pgfplots}
\usepackage{microtype}
\usepackage{inconsolata}
\usetikzlibrary{positioning}
\usepackage{arydshln}
\usepackage[T1]{fontenc}

\usepackage[utf8]{inputenc}
\usepackage{caption}
\usepackage{subcaption}
\usepackage{xspace}
\usepackage{xcolor}

\definecolor{purple}{rgb}{0.5,0,1}
\definecolor{dcyan}{rgb}{0.2,0.6,0.5}
\definecolor{light-gray}{gray}{0.95} 
\definecolor{darkgreen}{RGB}{0,140,0}
\definecolor{darkred}{RGB}{200,0,0}
\definecolor{lightgreen}{RGB}{189,252,192}
\definecolor{lightred}{RGB}{255,205,212}
\definecolor{lightyellow}{RGB}{255,240,160}
\definecolor{lightblue}{RGB}{195,221,255}
\definecolor{lightpurple}{RGB}{232,209,255}

\newcommand{\entaillabel}[1]{\textcolor{mdgreen}{\textsc{#1}}\xspace}

\usepackage{xcolor}

\definecolor{orange}{rgb}{1,0.5,0}
\definecolor{mdgreen}{rgb}{0.05,0.6,0.05}
\definecolor{mdblue}{rgb}{0,0,0.7}
\definecolor{dkblue}{rgb}{0,0,0.5}
\definecolor{dkgray}{rgb}{0.3,0.3,0.3}
\definecolor{slate}{rgb}{0.25,0.25,0.4}
\definecolor{gray}{rgb}{0.5,0.5,0.5}
\definecolor{ltgray}{rgb}{0.7,0.7,0.7}
\definecolor{purple}{rgb}{0.7,0,1.0}
\definecolor{lavender}{rgb}{0.65,0.55,1.0}

\usepackage{arydshln}

\usepackage{paralist}
\usepackage[para]{footmisc}
\usepackage{amsmath}
\usepackage{booktabs}

\newcommand{\entail}{\entaillabel{Entail}}

\newcommand{\contradict}{\entaillabel{Contradict}}
\newcommand{\neutral}{\entaillabel{Neutral}}
\newcommand{\nonneutral}{\entaillabel{Non-Neutral}}
\newcommand{\alphaOne}{$\alpha_1$\xspace}
\newcommand{\alphaTwo}{$\alpha_2$\xspace}
\newcommand{\alphaThree}{$\alpha_3$\xspace}

\newcommand{\datasetName}{{\sc InfoTabS}\xspace}
\newcommand{\ourdatasetName}{{\sc Auto-TNLI}\xspace}
\usepackage{microtype}

\usepackage{listings}

\usepackage{graphicx}
\usepackage{stfloats}
\usepackage{tabularx,ragged2e}
\usepackage{multirow}
\usepackage{paralist}
\usepackage[para]{footmisc}

\usepackage{color, colortbl}
\definecolor{LightCyan}{rgb}{0.88,1,1}

\definecolor{ecolor}{RGB}{44, 180, 44}
\definecolor{necolor}{RGB}{255,69,0}

\title{Realistic Data Augmentation Framework for Enhancing Tabular Reasoning}

\author {
    Dibyakanti Kumar\textsuperscript{\rm 1\thanks{Equal Contribution}},
    Vivek Gupta\textsuperscript{\rm {2*}\thanks{Corresponding Author}},
    Soumya Sharma\textsuperscript{\rm 3},
    Shuo Zhang\textsuperscript{\rm 4}
    \\\textsuperscript{\rm 1}IIT Guwahati;
     \textsuperscript{\rm 2}University of Utah;
     \textsuperscript{\rm 3}IIT Kharagpur;
     \textsuperscript{\rm 4}Bloomberg;
     \\
    dibyakan@iitg.ac.in; vgupta@cs.utah.edu; soumyasharma20@gmail.com; szhang611@bloomberg.net
}
\begin{document}












\maketitle
\begin{abstract}
Existing approaches to constructing training data for Natural Language Inference (NLI) tasks, such as for semi-structured table reasoning, are either via crowdsourcing or fully automatic methods. However, the former is expensive and time-consuming and thus limits scale, and the latter often produces naive examples that may lack complex reasoning.
This paper develops a realistic semi-automated framework for data augmentation for tabular inference.
Instead of manually generating a hypothesis for each table, our methodology generates hypothesis templates transferable to similar tables. In addition, our framework entails the creation of rational counterfactual tables based on human written logical constraints and premise paraphrasing.
For our case study, we use the \datasetName \cite{gupta-etal-2020-infotabs}, which is an entity-centric tabular inference dataset. We observed that our framework could generate human-like tabular inference examples, which could benefit training data augmentation, especially in the scenario with limited supervision.

\end{abstract}
\section{Introduction}

Natural Language Inference (NLI) is a Natural Language Processing task of determining if a hypothesis is entailed or contradicted given a premise or is unrelated to it \citep{dagan2013recognizing}. 
The NLI task has been extended for tabular data where it takes tables as the premise instead of sentences, namely tabular inference task. Two popular human-curated datasets for tabular reasoning, TABFACT \citep{chen2019tabfact} and \datasetName \citep{gupta-etal-2020-infotabs} datasets, have enhanced recent research in this area. 

However, human-generated datasets are limited in scale and thus insufficient for learning with large language models ~\cite[e.g.,][]{devlin2019bert,liu2019roberta}. Since curating these datasets requires expertise, huge annotation time, and expense, they cannot be scaled. 
Furthermore, it has been shown that these datasets suffer from annotation bias and spurious correlation problem ~\cite[e.g.,][]{poliak2018hypothesis,gururangan2018annotation,geva-annotator}. 
In contrast, automatically generated data lacks diversity and have naive reasoning aspects. Recently, use of large language generation model ~\cite[e.g.,][]{radford2018improving,lewis-etal-2020-bart,raffel2020exploring} is also proposed for data generation ~\cite[e.g.,][]{zhao2021lmturk,ouyang2022training,naturalinstructionsv1}. Despite substantial improvement, these generation approaches still lack factuality, i.e., suffer hallucination, have poor facts coverage, and also suffer from token repetition (refer to Appendix \S\ref{sec:appendi_auto-gen} analysis). Recently, \citet{chen-etal-2020-logical} shows that automatic tabular NLG frameworks cannot produce logical statements and provide only surface reasoning.

To address the above shortcomings, we propose a semi-automatic framework that exploits the patterns in tabular structure for hypothesis generation. Specifically, this framework generates hypothesis templates transferable to similar tables since tables with similar categories, e.g., two athlete tables in Wikipedia, will share many common attributes. In Table \ref{tab:example} the premise table key attributes such as ``Born'', ``Died'', ``Children'' will soon be shared across other tables from the ``Person'' category. One can generate a template for tables in the Person category, such as <Person$\_$Name> died before/after <Died:Year>. This template could be used to generate sentences as shown in Table \ref{tab:example} hypothesis H1 and H1$^C$. Furthermore, humans can utilize cell types (e.g., Date, Boolean) for generation templates. Recently, it has been shown that training on counterfactual data enhances model robustness \citep{Mller2021TAPASAS, wang2021robustness, rajagopal2022counterfactual}. Therefore, we also utilize the overlapping key pattern to create counterfactual tables. The complexity and diversity of the templates can be enforced via human annotators. Additionally, one can further enhance the diversity by automatic/manual paraphrasing \citep{dagan2013recognizing} of the template or generated sentences.  

\begin{table*}[ht]
\small
\centering
\begin{tabularx}{\textwidth}{@{} X X X | X X X @{}}
 \hline
 \multicolumn{3}{c|}{\textbf{Janet Leigh (Original)}} & \multicolumn{3}{c}{\textbf{Janet Leigh (Counter-Factual)}} \\ \hline
 \multicolumn{1}{l}{\textbf{Born}} & \multicolumn{1}{r}{July 6, 1927} & & \multicolumn{1}{l}{\textbf{Born}} & \multicolumn{1}{r}{July 6, 1927} &\\  
 \multicolumn{1}{l}{\textbf{Died}} & \multicolumn{1}{r}{October 3, 2004} & &  \multicolumn{1}{l}{\textbf{Died}} & \multicolumn{1}{r}{January 13, 1994} &\\  
 \multicolumn{1}{l}{\textbf{Children}} & \multicolumn{1}{r}{Kelly Curtis; Jamie Lee Curtis} & & \multicolumn{1}{l}{\textbf{Children}} & \multicolumn{1}{r}{Kelly Curtis} &\\
 \multicolumn{1}{l}{\textbf{Alma Mater}} & \multicolumn{1}{r}{Stanford University} & &  \multicolumn{1}{l}{\textbf{Alma Mater}} & \multicolumn{1}{r}{University of California} & \\
 \multicolumn{1}{l}{\textbf{Occupation}} & \multicolumn{1}{r}{None} & &  \multicolumn{1}{l}{\textbf{Occupation}} & \multicolumn{1}{r}{Scientist} & \\ \hline
 \multicolumn{2}{l}{H1: Janet Leigh was born before 1940.} & \multicolumn{1}{r|}{\textcolor{ecolor}{E}} & \multicolumn{2}{l}{H$1^C$: Janet Leigh was born after 1915.} & \multicolumn{1}{r}{\textcolor{ecolor}{E}}\\
 \multicolumn{2}{l}{H2: The age of Janet Leigh is more than 70.} & \multicolumn{1}{r|}{\textcolor{ecolor}{E}} & \multicolumn{2}{l}{H$2^C$: The age of Janet Leigh is more than 70.} & \multicolumn{1}{r}{\textcolor{necolor}{C}}\\
 \multicolumn{2}{l}{H3: Janet Leigh has 1 children} & \multicolumn{1}{r|}{\textcolor{necolor}{C}} & \multicolumn{2}{l}{H$3^C$: Janet Leigh has more than 2 children.} & \multicolumn{1}{r}{\textcolor{necolor}{C}}\\
 \multicolumn{2}{l}{H4: Janet Leigh graduated from Stanford University } & \multicolumn{1}{r|}{\textcolor{ecolor}{E}} & \multicolumn{2}{l}{H4$^C$: Janet Leigh graduated from Stanford University} & \multicolumn{1}{r}{\textcolor{necolor}{C}}\\
 \hline
\end{tabularx}
\vspace{-0.5em}
\caption{A example of an original and counterfactual table from the "Person" category. Here, we illustrate how multiple operations can be used to alter different keys. In addition, we have shown how the labels (\textcolor{ecolor}{E - Entail}, \textcolor{necolor}{C - Contradict}) for a specific hypothesis can alter. In the ``Janet Leigh'' example table, the first column represents the keys (e.g. Born; Died) and the second column has the relevant values (e.g. July 6,1927; October 3, 2004 etc).}
\vspace{-1.0em}
\label{tab:example}
\end{table*}

To show the effectiveness of our proposed framework, we conduct a case study with \datasetName dataset. \datasetName is an entity-centric dataset for tabular inference, as shown in example Table \ref{tab:example}. We extend the \datasetName data (25K table-hypothesis pair) by creating \ourdatasetName, which consists of  1,478,662 table-hypothesis pairs derived from 660 human written templates based on 134 unique table keys from 10,182 tables. 
For experiments, we utilize \ourdatasetName in three ways (a.) as a standalone tabular inference dataset for benchmarking, (b.) as a potential augmentation dataset to enhance tabular reasoning on \datasetName, i.e., the human-created data (c.) as evaluation set to assess model reasoning ability. We show that \ourdatasetName is an effective data for benchmarking and data augmentation, especially in a limited supervision setting. Thus, this semi-automatic generation methodology has the potential to provide the best of both worlds (automatic and human generation). 

To summarize, we make the following contributions in this paper:

\begin{itemize}
    \item We propose a semi-automatic framework that exploits the patterns in tabular structure for hypothesis generation.
    \item We apply this framework to extend the \datasetName ~\cite{gupta-etal-2020-infotabs} dataset and create a large-scale human-like synthetic data \ourdatasetName that contains counterfactual entity-based tables. 
    \item We conduct intensive experiments using \ourdatasetName and demonstrate it helps benchmark and data augmentation, especially in a limited supervision setting.
\end{itemize}

The dataset and associated scripts, are available at
\url{https://autotnli.github.io}.

\begin{table*}[!b]
\small
\centering
\begin{tabular}{lccccccccccc}
\hline
\textbf{Train-Data} & \textbf{City} & \textbf{Album} & \textbf{Person} & \textbf{Movie} & \bf Book & \bf F\&D & \bf Org & \bf Paint & \bf Fest & \bf S\&E & \bf Univ \\
\hline
\bf Orig & 78.32 & 67.81 & 92.45 & 97.12 & 96.31 & 92.27 & 92.44 & 98.93 & 87.44 & 82.53 & 85.59 \\
\bf Orig +Count & 61.89 & \bf 68.26 & 94.45 & 98.67 & \bf 98.72 & 97.04 & 96.46 & 99.56 & \bf 93.73 & \bf 95.68 & 93.02 \\
\bf MNLI +Orig & \bf 78.6 & 68.12 & 92.89 & 97.74 & 97.21 & 93.19 & 93.06 & 99.36 & 88.12 & 84.18 & 87.03 \\
\bf MNLI +Orig +Count & 62.32 & 68.01 & \bf 94.54 & \bf 99.01 & 98.46 & \bf 97.47 & \bf 96.8 & \bf 99.63 & 93.66 & 95.08 & \bf 93.56 \\
\hline
\end{tabular}
\vspace{-0.5em}
\caption{ Category-wise results for \ourdatasetName (\textbf{F\&D}- Food \& Drinks, \textbf{S\&E} - Sports \& Events)}
\label{fig:Category}
\vspace{-0.5em}
\end{table*}

\section{Proposed Framework}
\label{FrameworkSection}

Our framework includes four main components: \begin{inparaenum}[(a.)] \item Hypothesis Template Creation, \item Rational Counterfactual Table Creation, \item Paraphrasing of Premise Tables, and \item Automatic Table-Hypothesis Generation. \end{inparaenum}.


\begin{figure*}[ht]
    \centering
    \includegraphics[width=0.80\textwidth]{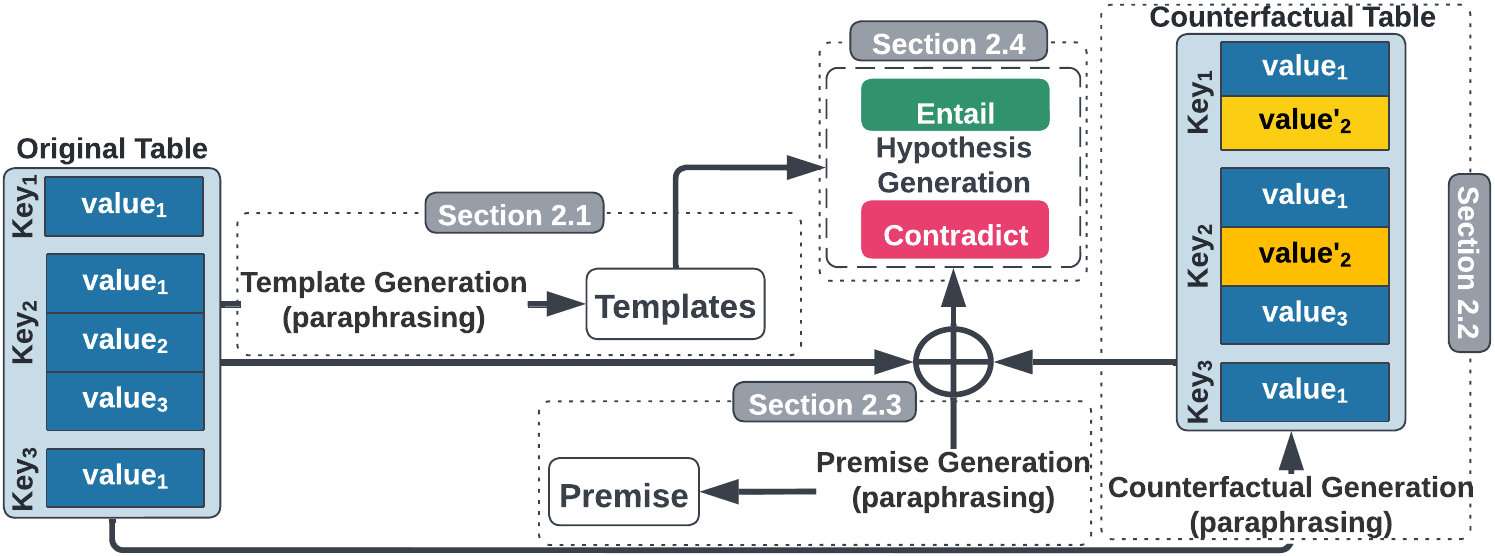}
    \vspace{-0.5em}
    \caption{\small Our Proposed Framework. \textbf{yellow} represents modified values in the counterfactual tables.}
    \label{fig:fig3}
    \vspace{-0.5em}
\end{figure*}


\begin{table*}[btp]
\small
\centering
\begin{tabular}{lcll}
 \toprule
 \bf Reasoning & \bf Category &\bf  Template-Rules &\bf Table-Constraints \\
 \midrule
 \multirow{2}{*}{\textbf{Temporal}} & \multirow{2}{*}{Person} & \multicolumn{1}{l}{\textit{<Person>} was born in a leap year.} & \multirow{1}{*}{Born Date $\leq$} \\
 & & \multicolumn{1}{l}{\textit{<Person>} died before/after \textit{<Died:Year>}} & \multirow{1}{*}{Death Date}\\ 
 \multirow{2}{*}{\textbf{Numerical}} & \multirow{2}{*}{Movie} & \multicolumn{1}{l}{\textit{<Movie>} was a "hit \lstinline{if}  \textit{<Box Office>} $-$ {\textit{<Budget>} \lstinline{else} flop"}} & \multirow{2}{*}{Budget $\geq$ $0$} \\
 & & \multicolumn{1}{l}{\textit{<Movie>} had a Box Office collection of \textit{<BoxOffice>}} & \\

  \multirow{2}{*}{\textbf{Spatial}} & \multirow{2}{*}{Movie} & \multicolumn{1}{l}{\textit{<Movie>} was released in \textit{<Release1:Loc>, "\textit{X}" months before/after}} & \multirow{1}{*}{Release1:Location $\neq$} \\
 & &  \multicolumn{1}{l}{ \textit{<Release2:Location>} } & \multicolumn{1}{l}{Release2:Location} \\
 
 
  \multirow{2}{*}{\textbf{KCS}} & \multirow{2}{*}{City} & \multicolumn{1}{l}{The governing of \textit{<City>} is supervised by \textit{<Mayor>}} & \multirow{1}{*}{Lowest Elevation $\leq$ } \\
 & & \multicolumn{1}{l}{\textit{<Mayor>} is an important local leader of \textit{<City>}} & \multirow{1}{*}{Highest Elevation} \\
 \bottomrule
\end{tabular}
\vspace{-0.5em}

\caption{Rules and Constraints are classified into specific areas of reasoning, as indicated in the table. A few examples of rules and constraints have been provided for each category. \textit{<Died:Year>} indicates that the year value is extracted from \textit{<Died>} , whereas  \textit{<Release1:Location>} indicates that the location is extracted from a single key-value pair in \textit{<Release>}. KCS denote knowledge and common sense reasoning in this context.} 
\vspace{-1.5em}

\label{tab:rules&constraints}
\end{table*}

\subsection{Hypothesis Template Creation}
\label{sec:tempgeneration}
For a particular category of tables (e.g., \emph{movie}), the row attributes (i.e. keys) are mostly overlapping across all tables (e.g., \emph{Length}, \emph{Producer}, \emph{Director}, and others). 
Therefore, this consistency across table benefits in writing table category specific \textbf{key-based rules} to create logical hypothesis sentences. We create such key-based rules for the following reasoning types: 
\begin{inparaenum}[(a.)] \item Temporal Reasoning, \item Numerical Reasoning, \item Spatial Reasoning, \item Common Sense Reasoning. \end{inparaenum} Table \ref{tab:rules&constraints} provide examples of logical rules used to create templates. We denote the category of a table as \textbf{\textit{Category}} and the table row keys of as \textit{<Key>}. In addition, each template is paraphrased to enhance lexical diversity.

Frequently, these key-based reasoning rules generalize effectively across several categories. For example, the temporal reasoning rule based on the date-time type could be minimally modified  to work for  \textit{<Release Date>} of category \textbf{\textit{Movies}} tables, as well as the \textit{<Established Date>} of category \textbf{\textit{University}} tables, in addition to the \textit{<Born>} of category \textbf{\textit{Person}} in Table \ref{tab:rules&constraints}. Additionally, reasoning rules can be expanded to incorporate multi-row entities from the same table's data, as illustrated in Table  \ref{tab:rules&constraints} for the numerical reasoning type. Other examples for the same are "The elevation range of \textit{<City>} is \textit{<HighestElevation>} $-$ \textit{<LowestElevation>}" for category \textbf{\textit{City}} table, "\textit{<SportName>} was held at \textit{<location>} on \textit{<date>}" for \textbf{\textit{Sports}} category.

\subsection{Rational Counterfactual Table Creation}
\label{sec:counterfacttable}
We also construct counterfactual tables, as illustrated in Table \ref{tab:example}, in which the values corresponding to the original table's keys are altered. This counterfactual table contains non-factual unreal information but is consistent, i.e., the table facts are not self contradictory. Language models trained on such counterfactual instances exhibit greater robustness \citep{Mller2021TAPASAS, wang2021robustness, rajagopal2022counterfactual,table-probing} and prevent the model from over-fitting its pre-learned knowledge. Benefiting model in grounding and examining the premise evidence as opposed to employing spurious correlation. To create counterfactual  table, we modify an original table with $k$ keys. For a given category, these $k$ keys constitute a subset of the $n$ possible unique keys ($n >= k$) for that category. 

To construct a counterfactual table, we modify the original table in one or more of the following ways: \begin{inparaenum}[(a.)] \item keep the row as it without any change,  \item adding new value to an existing key, \item substituting the existing key-value with counter-factual data, \item deleting a particular key-value pair from the table, \item and add a missing new keys (i.e. a key from ($n-k$) ), \item and adding a missing key row to the table. \end{inparaenum} For creating counterfactual tables, for each row of existing, a subset of operation is selected at a random each with a pre-decided  probability $p$ (a hyper-parameter).





 While creating these tables, we impose an essential key-specific constraints to ensure logical rational in the generated sentences. E.g. in the example Table \ref{tab:example}, for the counterfactual table of \emph{Janet Leigh (Counterfactual)}, the \textit{<Born>} is kept similar to original of \emph{Janet Leigh (Original)}, whereas \textit{<Died>} has been substituted for another \emph{Person} table, while ensuring the constraint {\sc Born Date} < {\sc Death Date} i.e. Jan 13, 1994 (Died Date of Counterfactual Table) is after July 6, 1927 (Born Date of Counterfactual Table)). Without the following the constraint that {\sc Born Date} < {\sc Death Date}, the table with become rationally incorrect or self contradictory.

\subsection{Paraphrasing of Premise Tables}
\label{sec:paraphrasing}
Lack of linguistic variety is a significant concern with grammar-based data generating methods. Therefore, we employ both automated and human paraphrase of premise tables to address diversity problem.
For each key for of a given category, we create at least three to five simple paraphrased sentences of the key-specific template. E.g. for \textit{<Alma Mater>} from \textbf{\textit{Person}}, possible paraphrases can be "\textit{<PersonName>} earned his degree from \textit{<AlmaMater>}", "\textit{<PersonName>} is a graduate of \textit{<AlmaMater>}", and "\textit{<AlmaMater>} is a alma mater of \textit{<PersonName>}". We observe that paraphrasing considerably increases the variability across instances.

\subsection{Automatic Table-Hypothesis Generation}
\label{sec:generation}
Once the templates are constructed as discussed in \S \ref{sec:tempgeneration}, they can be used to automatically fill in the blanks from the entries of the considered tables and create logically rational hypothesis sentences. To create contradictory sentences, we randomly select a value from a collection of key values shared by all tables to fill in the blanks. This replacement ensures that the key-specific constraints, such as the key-value type, are adhered to. Furthermore, we ensure that similar template with minimal token alteration is used to create entail contradict pair. This way of creating entail and contradiction statement pairs with lexically overlapping tokens ensure that, future model trained on such data won't adhere spurious correlation from the tabular NLI data i.e. minimising the hypothesis bias problem \citep{poliak2018hypothesis}. For example, for movie "Ironman" movie with rows "Budget:\$140 million" and "Box-office:\$585.8 million", using the template {\textit{<Movie>} was a "hit \lstinline{if}  \textit{<Box Office>} $-$ {\textit{<Budget>} \lstinline{else} flop"}} from example Table ~\ref{tab:rules&constraints}, one can generate hypothesis \textit{entail}: "The movie Ironman was a hit" and \textit{contradict}:  "The movie Ironman was a flop".


\section{The \texttt{\ourdatasetName} Dataset}

\label{sec:autotnlidataset}
We apply our framework as described in \S\ref{FrameworkSection} on an entity specific tabular inference dataset \datasetName to construct \ourdatasetName. 
\datasetName \citep{gupta-etal-2020-infotabs} consists of pairs of NLI instances:  a hypothesis statement grounded and inferred on premise table is extracted from Wikipedia Infobox table across multiple diverse categories. We construct the \ourdatasetName dataset from a subset of the \datasetName dataset ($11$ out of $13$ total categories), which includes the original table plus five counterfactual tables corresponding to each original table, for a total of $10,182$ tables. We retrieve $134$ keys and $660$ templates, which we utilize to generate $1,478,662$ sentences. However, unlike \datasetName, which contains $3$ labels, \entail, \contradict and \neutral, \ourdatasetName contains only two labels \entail and \contradict.

\begin{table}[!htbp]
\small
\centering
\begin{tabular}{lc}
\hline
\textbf{Statistic Metric} & \multicolumn{1}{r}{\textbf{Numbers}} \\
\hline
Number of Unique Keys & \multicolumn{1}{r}{134} \\
Average number of keys per table & \multicolumn{1}{r}{12.63} \\
Average number of sentences per table & \multicolumn{1}{r}{164.51} \\
\hline
\end{tabular}
\vspace{-0.5em}
\caption{\ourdatasetName~ Statistics.}
\vspace{-2.0em}
\label{tab:Stats1}
\end{table}

As previously reported in the original \datasetName paper by \citet{gupta-etal-2020-infotabs}, annotators are biased towards specific keys over others. For example, for the category \textbf{\textit{Company}}, annotators would create more sentences for the key \textit{<Founded by>} than for the key \textit{<Website>}, resulting in an inherent hypothesis bias in the dataset. While creating the templates for \ourdatasetName, we ensure that each key has a minimum of two hypotheses and a minimum of three ($>3$) premise paraphrases, which helps mitigate hypothesis bias. To address the inference class imbalance labeling issue, we construct approximately 1:1 \entail to \contradict the hypothesis.



We observe that most additional human labor required to build such sentences is spent on the set of key-specific rules and constraints that ensure the sentences are grammatically accurate. The counter-factual tabular data is logically consistent, i.e., not self-contradictory. Table \ref{tab:Stats1} details the number of unique keys, the minimum/maximum/average number of keys, and the total number of sentences per table in \ourdatasetName. As can be observed, the system generates a large amount of \ourdatasetName data compared to limited \datasetName while using only a few human-constructed templates with key-specific rules and constraints. 


We have chosen \datasetName as it has three evaluation sets \alphaOne, \alphaTwo, and \alphaThree, in addition to the regular training and development sets. The \alphaOne set is lexically and topic-wise similar to the train set, and in \alphaTwo the hypothesis is lexically adversarial to the train set. And in \alphaThree the tables are from topics not in the train set. Moreover, it has multiple reasoning types such as multi-row reasoning, entity type, negation, knowledge \& common sense, etc. \datasetName has all three labels \entail, \neutral, and \contradict compared to just two labels in other datasets such as TABFACT. 

\paragraph{Human Verification}
To evaluate the quality and correctness of our data, we requested one of our human annotators (expert NLP Ph.D. Grad student) to assign a label to the generated hypothesis and select a score from 1 to 5 for the grammar and complexity of the sentences. The grammar score reflects how meaningful and lexically accurate the data is, and the complexity score indicates how difficult it is to label the hypothesis correctly. This was done for about 1300 premise-hypothesis pairs from \ourdatasetName.
\begin{table}[!h]
\small
\centering
\begin{tabular}{l|c}
\toprule
\textbf{Statistic Metric} &\bf Numbers \\
\midrule
Percentage of correct labels ($\%$) & 99.4 \\
Average Grammar score (1-5) & 4.89 \\
Average Complexity score (1-5) & 3.64 \\
\bottomrule
\end{tabular}
\vspace{-0.5em}
\caption{Human verification Statistics.}
\label{tab:Human_verification}
\vspace{-1.0em}
\end{table}
\newline
\textbf{Analysis}: As observed in Table \ref{tab:Human_verification}, humans marked 99.5\% of the data as correctly labeled and gave an average score of about 4.89 out of 5 for the grammatical accuracy of the sentences. The sentences in this data also received an average complexity score of 3.64 out of 5.

\section{Experiments and Analysis}

Overall, we address the following two research questions through our experiments:

\vspace{0.5em}
\noindent \textbf{RQ1:} (a) Taking \ourdatasetName as an evaluation set, how challenging is the TNLI task? (b) If fine-tuning on \ourdatasetName beneficial? 
\vspace{0.5em}

\noindent \textbf{RQ2:} (a) Is it beneficial to use \ourdatasetName as data augmentation for the TNLI task? (b) If so, will it also be useful in little supervision scenario?

\begin{table*}[t]
\small
\centering
\begin{tabular}{ll|cccccc}
\toprule
\textbf{Training} & \textbf{Augmentation Strategy} & \textbf{Cat-Ran} & \textbf{Cross-Cat} & \textbf{Key} & \textbf{NoPara} & \textbf{Cross-Para} & \textbf{Entity}\\
\midrule
\multirow{5}{5em}{\small \sc w/o \ourdatasetName}& w/o finetuning & 50.00 & 49.64 & 50.17 & 49.77 & 49.75 & 49.78 \\
& \datasetName & 66.17 & 63.86 & \bf 65.41 & 65.15 & 65.12 & 63.66 \\
& MNLI & 67.15 & 64.95 & 64.79 & 65.33 & 65.33 & 62.2 \\
& MNLI +\datasetName & \textbf{69.28} & \bf 65.9 & 65.25 & \bf{66.41} & \bf{66.39} & \bf{65.02} \\
\midrule
\multirow{4}{4em}{\small \sc w \ourdatasetName } & Hypothesis-Only & 53.74 & 55.1 & 58.82 & 66.47 & 66.86 & 56.36 \\
& \ourdatasetName & 78.74 & 77.94 & 82.39 & 90.06 & 89.38 & 74.94 \\
& MNLI +\ourdatasetName &\bf 83.82 & 78.95 & 84.71 & \bf 91.17 & \bf 90.57 & \bf 77.66 \\
& MNLI +\datasetName +\ourdatasetName & 83.62 & \bf 80.78 & \bf 85.23 & 90.98 & 90.03 & 77.19 \\
\bottomrule
\end{tabular}
\vspace{-0.5em}
\caption{Accuracy with RoBERTa$_{\text{BASE}}$ model across several evaluation splits with / without fine-tuning on \ourdatasetName. \textbf{bold} - represents max across rows i.e. best train/augmentation setting.}
\label{fig:RQ1}
\vspace{-1.5em}
\end{table*}

\paragraph{Experiment Settings.} We use RoBERTa$_{\text{BASE}}$ \citep{liu:19} ($12$-layer, $768$-hidden, $12$-heads, $125$M parameters) and ALBERT$_{\text{BASE}}$ \citep{lan2019albert} ($ 12$-layer, $768$-hidden, $12$-heads, $12$M parameters) as our model for all of our experiments \footnote{Due to the large scale of the \ourdatasetName data, we favour BASE over LARGE models for conducting efficient experiments.}. \citet{neeraja-etal-2021-infotabskg} shows data augmentation techniques that uses MNLI data for pre-training acts as implicit knowledge and enhances the model performance for \datasetName. Therefore, we also explore implicit knowledge addition via data augmentation. In particular, we explored the following models: \begin{inparaenum}[(a)] \item RoBERTa$_{\text{BASE}}$ fine-tuned using the \ourdatasetName dataset \item RoBERTa$_{\text{BASE}}$, fine-tuned on the MNLI dataset and the \ourdatasetName dataset (MNLI + \ourdatasetName). \end{inparaenum} Additionally, we also explore performance with RoBERTa$_{\text{BASE}}$ model fine-tuned sequential on all three MNLI, \ourdatasetName and \datasetName dataset. Due to limited space, we report all ALBERT \footnote{Experiments on the development set showed that RoBERTa$_{\text{BASE}}$ outperforms other pre-trained language models. BERT$_{\text{BASE}}$ and ALBERT$_{\text{BASE}}$ reached an accuracy of 63\% and 70.4\% respectively} findings in Appendix \ref{sec:appendixALBERT}. 

\subsection{Using \ourdatasetName as TNLI dataset}
\label{sec:RQ1}
In this section, we assess how challenging our \ourdatasetName is compared to the \datasetName datasets (i.e., RQ1).

\paragraph{Data Splits.} We first construct several train-dev-test splits of ~\ourdatasetName~such that:  \begin{inparaenum}[(a)] \item splits have table from different domains (categories)\footnote{by table domain/categories we refer to table entity types e.g. "Person", "Album", and others.} \item splits have unique table row-keys, \item premises in splits are lexically diverse. \end{inparaenum} For the category-wise splits, we explore two ways \begin{inparaenum}[(a)] \item we divided categories randomly into train-dev-test. \item we construct the splits after doing a cross-category performance analysis (refer \S\ref{sec:appendix} in the Appendix). In the cross-category analysis, we get all premise-hypothesis pairs generated from tables in one category (for example \emph{person}) and train our model on this data. After this we test on premise-hypothesis pairs generated from all other categories (for example : \emph{city}, \emph{movie} etc.) one-by-one. We keep the difficult categories for the model to solve in the test set. 
This is accomplished by counting the number of times an category's accuracy falls below a specific threshold\footnote{We choose the threshold as 80\%.} and then selecting the entities with the highest frequency. We kept \emph{book}, \emph{paint}, \emph{sports \& events}, \emph{food \& drinks}, \emph{album} in train-set, \emph{person}, \emph{movie}, \emph{city} in dev-set and \emph{organization}, \emph{festival}, \emph{university} in test-set. 

\end{inparaenum} For key-wise split, we explore two approaches \begin{inparaenum}[(a)] \item we divide the keys randomly into train-dev-test. \item we decided splits based on the associated keys-values named entities type namely - \emph{person}, \emph{person type}, \emph{skill}, \emph{organization}, \emph{quantity}, \emph{date time}, \emph{location}, \emph{event}, \emph{url}, \emph{product} after cross-entity performance analysis.\end{inparaenum}. Similar to cross-category analysis above, here we get all premise-hypothesis pairs corresponding to keys in a single entity, for example let's say we choose the entity \emph{person} and it includes the keys \emph{written by}, \emph{mayor}, \emph{president} etc. then we get all premise-hypothesis pairs corresponding to these keys and train on them. After this we test on premise-hypothesis pairs corresponding to all other entities (for example : \emph{persontype}, \emph{skill}) one-by-one. We select the entities that are challenging for the model in the test set. This is accomplished by counting the number of times an entity's accuracy falls below a specific threshold$^4$ and then selecting the entities with the highest frequency. We kept the \emph{url}, \emph{event}, \emph{person type}, \emph{skill}, \emph{product} in train-set, \emph{quantity}, \emph{other}, \emph{person} in dev-set and \emph{date time}, \emph{organization}, \emph{location} in test-set.  

Finally, for the lexical diversity, we split via paraphrasing premise. Here too, we explore two different strategies \begin{inparaenum}[(a)] \item premises in train, dev, and test are not paraphrased, i.e., have similar templates. \item premises in train, dev, and test are lexically paraphrased i.e. have distinct templates.\end{inparaenum}
\paragraph{Using \ourdatasetName only for Evaluation (RQ1a):}
We first explore how challenging is \ourdatasetName is used as an evaluation benchmark dataset. To explore this, we compare the performance of pre-trained RoBERTa$_{\text{BASE}}$ model in four distinct settings, as follows  \begin{inparaenum}[(a.)]
\item without (w/o) fine-tuning,
\item fine-tuned with \datasetName,
\item fine-tuned with MNLI,
\item fine-tuned over both MNLI and \datasetName in order
\end{inparaenum} and and evaluate it on \ourdatasetName test-sets splits. For finetuning on MNLI and \datasetName dataset, we only consider the \entail and \contradict while excluding the \neutral label instances for training purposes.

\textit{Analysis.} Table \ref{fig:RQ1} shows a comparison of accuracy across all augmentation settings. The best is obtained when using both MNLI and \datasetName for training. In the cases where we have used some fine-tuning with MNLI or \datasetName we observed an average accuracy of 67.5\%. Comparing this with zero-shot accuracy for \datasetName where we observed accuracy of 58.9\%, we can see that semi-automatically generated data is still challenging. 


\paragraph{Using \ourdatasetName for both Training and Evaluation (RQ1b):}
Next, we explore if providing supervision improves the performance on the \ourdatasetName evaluation sets. To explore this, we compare the performance of pre-trained RoBERTa$_{\text{BASE}}$ model in two distinct settings, where we fine-tune on train set \begin{inparaenum}[(a.)]
\item of \ourdatasetName,
\item of both MNLI and \ourdatasetName in order
\end{inparaenum} and evaluate on \ourdatasetName test-sets. Here too, we exclude the \neutral label instances from MNLI.

\textit{Analysis.} Table \ref{fig:RQ1} shows a performance (accuracy) comparison across all augmentation settings. For all splits except paraphrasing, RoBERTa$_{\text{BASE}}$ achieves an average 80\% accuracy. It shows that our semi-automated dataset \ourdatasetName is as challenging as \datasetName \citep{gupta-etal-2020-infotabs}, which has an average accuracy of 70\% across all splits and is manually human-generated and is one-tenth the size of \ourdatasetName. Pre-finetuning with MNLI as augmented data (i.e., implicit knowledge) only improves the performance by 2$\%$. Identical findings were also seen with ALBERT$_{\text{BASE}}$ model, c.f. Appendix \ref{sec:appendixALBERT} Table \ref{tab:albert_RQ1}.


\subsection{Using \ourdatasetName for Data Augmentation}
\label{sec:Data_aug}
\begin{figure}[t]
    \centering
    \includegraphics[width=0.36\textwidth]{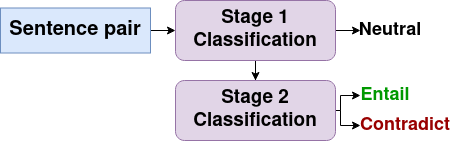}
    \caption{Two stage classification approach.} 
    \vspace{-0.5em}
    \label{fig:twostage-approach}
    \vspace{-1.0em}
\end{figure}

We explore if \ourdatasetName can be used as an augmentation dataset for \datasetName (i.e. RQ2). Since \datasetName include all three \entail, \neutral and \contradict labels, whereas \ourdatasetName include only \entail and \contradict labels, we explore the inference task as a two-stage classification problem. In first stage, we train a RoBERTa$_{\text{BASE}}$ classification model to predicts whether a hypothesis is \neutral vs \nonneutral (either \entail or \contradict). 
In second stage, we fine-tune a separate RoBERTa$_{\text{BASE}}$ model to further classify the \nonneutral prediction instances from stage one into \entail or \contradict label. Figure \ref{fig:twostage-approach} illustrate the two-stage classification approach.

\paragraph{Comparison Models.}
For first-stage we consider two training strategies:
\begin{inparaenum}[(a.)]
\item only train on \datasetName,
\item pre-finetune on both MNLI followed by training on \datasetName.
\end{inparaenum} We consider multiple data augmentation technique for second stage training where we augment \begin{inparaenum}[(a.)] 
\item \textbf{Orig:} the \ourdatasetName without counterfactual table instances,
\item \textbf{Orig +Count:}  \ourdatasetName including counterfactual table instances\footnote{We take five counterfactual tables for each original table.},
\item \textbf{MNLI +Orig:} both MNLI and \ourdatasetName without counterfactual table instances,
\item \textbf{MNLI +Orig +Count:} both MNLI and \ourdatasetName including counterfactual table instances.
\end{inparaenum} Additionally, we compare all above methods with (e.) \textbf{No Aug} i.e. the approach where we do not augment any additional data.

\paragraph{Evaluation Set.} We utilize the \datasetName test sets, which include all three inference labels for evaluation. In addition to standard development and a test split (\alphaOne), \datasetName also has two adversarial test splits, namely \alphaTwo and \alphaThree. 
E.g. in the example Table \ref{tab:example} if hypothesis sentence \textit{Janet Leigh was born \emph{before} 1940} is \entail, then in \alphaTwo after perturbation the instance became \textit{Janet Leigh was born \emph{after} 1940} with label as \contradict. The test set \alphaThree is a zero-shot evaluation set consisting of premise tables from different domains with minimal key overlaps with the training set premise tables. To better handle \alphaTwo and \alphaThree test-sets, we include a counterfactual table and hypothesis in \ourdatasetName.

\paragraph{Supervision Scenarios.} We analyse the effect of using  \ourdatasetName as augmentation data for \datasetName in two setting \begin{inparaenum}[(a)] \item \textbf{Complete Supervision} where we use complete \datasetName training set for final fine-tuning \item \textbf{Limited Supervision} where we use limited \datasetName supervision for second stages. We explore using 0$\%$ (i.e. no fine-tune), 5$\%$, 15$\%$ and 25$\%$ of \datasetName training set for final fine-tuning. \end{inparaenum} 
\newline

\begin{table}[!ht]
\vspace{-1.0em}
\small
\centering
\begin{tabular}{l|ccccc}
\hline
& \multicolumn{5}{|c}{\textbf{Stage 2: Entail  vs Contradict}} \\
\hline
 &   &  & & \textbf{MNLI} & \textbf{MNLI} \\
\textbf{Split} &\textbf{No} & \textbf{Orig} & \textbf{Orig} & \textbf{+Orig} & \textbf{+Orig} \\
 &\textbf{Aug} & & \textbf{+Count} & & \textbf{+Count}\\
\hline
& \multicolumn{5}{c}{\bf Stage 1: \datasetName} \\
\hline
 dev & 71.06  & 70.72 & 71.39 & \textbf{72.28} & 72.22 \\
\alphaOne & 67.72 & 67.56 & 69.33 & 68.78 & \textbf{69.89} \\
\alphaTwo & 59.11 & 59.22 & 58.94 & 59.5 & \textbf{61.28} \\
\alphaThree & 56.94 & 56.94 & 58.17 & 58.33 & \textbf{58.61} \\
\hline
& \multicolumn{5}{c}{\bf Stage 1: MNLI +\datasetName}\\
\hline
 dev & 70.67 & 70.89 & 71.44 & 72.56 & \textbf{72.67} \\
\alphaOne & 68.94 & 68.83 & 70.56 & 70.67 & \textbf{72.00} \\
\alphaTwo & 60.56 & 60.83 & 60.5 & 61.11 & \textbf{62.50} \\
\alphaThree & 58.44 & 57.72 & 59.11 & \textbf{60.06} & 59.94 \\
\hline
\end{tabular}
\vspace{-0.5em}
\caption{Accuracy of combine stage I i.e. \neutral vs \nonneutral and stage II i.e. \entail vs \contradict classifiers (RoBERTa$_{\text{BASE}}$) across several data augmentation settings. Here, for stage one we also explore pre-fine tuning on MNLI data. \textbf{bold} - represents max across columns i.e. the best augmentation setting.}
\label{tab:RQ2}
\vspace{-1.5em}
\end{table}

\paragraph{1. Complete \datasetName Supervision (RQ2a)}

Table \ref{tab:RQ2} shows a comparison of accuracy across all augmentation settings. 

In the \textbf{first case}, when the first stage is only trained on \datasetName, we observe an improvement of 1.6$\%$ and 1.2$\%$ percentage in \alphaOne and \alphaThree test-set through direct \ourdatasetName data augmentation base pre-finetuning (Orig+Count) in comparison with no augmentation i.e. direct \datasetName fine-tuning. We didn't see any substantial improvement in \alphaTwo performance. Fine-tuning with MNLI followed by \ourdatasetName (with counterfactual tables) further improve the performance by 0.6$\%$, 2.0$\%$, and 0.45$\%$ on \alphaOne, \alphaTwo, and \alphaThree respectively. 

For \textbf{second case}, when the first stage is trained on both MNLI, followed by \datasetName, we observe an improvement of 1.60$\%$ and 0.67$\%$ percentage in \alphaOne and \alphaThree test-set through direct \ourdatasetName data augmentation base pre-finetuning (Orig+Count) in comparison with no augmentation i.e. direct \datasetName fine-tuning. Here too, we didn't see any substantial improvement in \alphaTwo performance. Finetuning with MNLI followed by \ourdatasetName (with counterfactual tables) further improve the performance by 1.44$\%$, 1.94$\%$, and 0.83$\%$ on \alphaOne, \alphaTwo and \alphaThree respectively. Identical findings were also seen with ALBERT$_{\text{BASE}}$ model, c.f. Appendix \ref{sec:appendixALBERT} table \ref{tab:albert_RQ2}.


\paragraph{Ablation Analysis - Independent Stage-1 and Stage-2 Performance:} We also did an ablation study to access the performance of individual RoBERTa$_{\text{BASE}}$ models of both stages. Table \ref{fig:AblationStage1}, show the performance for stage one classifier i.e. \neutral vs \nonneutral. We observe that adding MNLI data for augmentation substantially improves the performance by 1.89$\%$, 2.28$\%$, and 2.05$\%$ for \alphaOne, \alphaTwo, and \alphaThree, respectively. 

\begin{table}[!htbp]
\small
\centering
\begin{tabular}{l|ccccc}
\hline
 &   &  & & \textbf{MNLI} & \textbf{MNLI} \\
\textbf{Split} &\textbf{No} & \textbf{Orig} & \textbf{Orig} & \textbf{+Orig} & \textbf{+Orig} \\
 &\textbf{Aug} & & \textbf{+Count} & & \textbf{+Count}\\
\hline
dev & 77.5 & 77.83 & 78.08 & \textbf{80.75} & 80.25 \\
\alphaOne & 73.58 & 73.83 & 76.33 & 76.5 & \textbf{79.00} \\
\alphaTwo & 56.92 & 57.42 & 56.92 & 58.42 & \textbf{60.25} \\
\alphaThree & 70.58 & 69.42 & 72 & \textbf{73.08} & 72.58 \\
\hline
\end{tabular}
\vspace{-0.5em}
\caption{Performance (accuracy) of stage two RoBERTa$_{\text{BASE}}$ (i.e. \entail vs \contradict) classifier across several data augmentation settings. \textbf{bold} same as Table \ref{tab:RQ2}. }
\label{fig:AblationStage2}
\vspace{-1.0em}
\end{table}

Table \ref{fig:AblationStage2} shows the comparison between all settings of stage-2. In stage-2 adding counterfactual tables improve the performance by 2.75\% and 1.42\% in \alphaTwo and \alphaThree respectively. We didn't see any substantial improvement in \alphaTwo performance. If we pre-finetune further with MNLI along with \ourdatasetName we further get an improvement of 5.42\%, 3.33\% and 2\% in \alphaOne, \alphaTwo, and \alphaThree respectively. Identical findings were also seen with ALBERT$_{\text{BASE}}$ model, c.f. Appendix \ref{sec:appendixALBERT} Table \ref{tab:albert_AblationStage1} and \ref{tab:albert_AblationStage2}.

\begin{table}[!htbp]
\small
\centering
\label{fig:Ablation-stage1} 
\begin{tabular}{l|cc}
\toprule
\textbf{Test-split} &\bf No Aug  & \textbf{MNLI} \\
\midrule
dev & 84.11 & \bf 84.50 \\
\alphaOne & 82.94 & \bf 84.83 \\
\alphaTwo & 85.33 & \bf 87.61 \\
\alphaThree & 73.17 & \bf 75.22 \\
\bottomrule
\end{tabular}
\vspace{-0.5em}
\caption{Performance (accuracy) of stage one RoBERTa$_{\text{BASE}}$ (i.e. \neutral vs \nonneutral) across several data augmentation settings. Here, No-Augmentation means \datasetName, and MNLI means MNLI + \datasetName. \textbf{bold} same as Table \ref{tab:RQ2}.}
\label{fig:AblationStage1}
\vspace{-1.0em}
\end{table}



\paragraph{2. Limited \datasetName Supervision (RQ2b)}

\begin{table}[t]
\small
\centering
\begin{tabular}{c|ccccc}
\hline
   &   &  & & \textbf{MNLI} & \textbf{MNLI} \\
\textbf{Tr(\%)} &\textbf{No} & \textbf{Orig} & \textbf{Orig} & \textbf{+Orig} & \textbf{+Orig} \\
 &\textbf{Aug} & & \textbf{+Count} & & \textbf{+Count}\\
\hline
& \multicolumn{5}{c}{\bf Development set}\\
\hline
0 & 50.25 & 59.58 & 52.58 & \textbf{62.67} & 60.75 \\
5 & 65.31 & 69.92 & 69.86 & 70.81 & \textbf{71.11} \\
15 & 69.47 & 71.69 & 73.61 & \textbf{75.28} & 74.42 \\
25 & 70.21 & 72.88 & 74.54 & \textbf{74.71} & 74.63 \\
\hline
& \multicolumn{5}{c}{\bf \alphaOne set}\\
\hline
 0 & 49.92 & 59.42 & 52.42 & 61.58 & \textbf{62.33} \\
5 & 65.75 & 69.08 & 68.89 & 70.72 & \textbf{70.92} \\
15 & 69.14 & 70.69 & 70.83 & 73.28 & \textbf{74.25} \\
25 & 69.75 & 72.38 & 73.75 & 74.5 & \textbf{75.13} \\
\hline
& \multicolumn{5}{c}{\bf \alphaTwo set}\\
\hline
0 & 50.17 & 59.00 & 59.75 & 61.17 & \textbf{61.67} \\
5 & 43.81 & 54.92 & 53.53 & 56.25 & \textbf{58.03} \\
15 & 47.31 & 54 & 53.03 & 56.89 & \textbf{57.42} \\
25 & 49.79 & 56.33 & 55.25 & \textbf{59} & 58.42 \\
\hline
& \multicolumn{5}{c}{\bf \alphaThree set}\\
\hline
0 & 49.42 & 59.25 & 56.33 & 64.67 & \textbf{63.92} \\
5  & 57.72 & 63.47 & 63.5 & 68.06 & \textbf{68.14} \\
15  & 64.42 & 65.69 & 68.47 & 70.03 & \textbf{71.11} \\
25  & 64.08 & 67.17 & 67.42 & 70.46 & \textbf{70.92} \\
\hline
\end{tabular}
\caption{Performance (accuracy) of RoBERTa$_{\text{BASE}}$ (i.e. \entail vs \contradict i.e. second stage) classifier with various data augmentation for limited supervision setting i.e. varying percentage of \datasetName training data. The average standard deviation across 3 runs is 1.36 with range 0.5$\%$ to 3.5$\%$. \textbf{bold} same as Table \ref{tab:RQ2}.}
\label{tab:limited_setting}
\vspace{-1.5em}
\end{table}

In this setting, we analyse the effect of limiting \datasetName supervision for the second stage i.e. \entail vs \contradict. We explore using 0$\%$ (i.e. no fine-tune), 5$\%$, 15$\%$ and 25$\%$ of \datasetName training set for fine-tuning. Table \ref{tab:limited_setting} shows the performance for every augmentation settings. The table report average result over three random samples from \ourdatasetName. We observe that augmenting with \ourdatasetName improve performance for all percentages. Furthermore, the improvement is much more substantial for lower than higher percentages. Here too, the best performance are obtained via fine-tuning with MNLI followed by \ourdatasetName for all percentages. In the Appendix Table \ref{tab:limited_setting_both_stages} and \ref{tab:limited_setting_both_stages_with_MNLI}, we present the combined stage performance on limited supervision both w and w/o MNLI pre-training. Refer the first stage results with limited supervision in Appendix Table \ref{tab:limited_setting_first_stage}. Appendix Figure \ref{fig:ConfusionMatrix} show consistency analysis. 
\section{Related Work}
\label{sec:related_work}
\paragraph{Tabular Reasoning.} There has been considerable work on solving NLP tasks on semi-structured tabular data, such as tabular NLI \citep{gupta-etal-2020-infotabs, chen2019tabfact,gupta-etal-2022-right}, question-answering task \citep[ and others]{zhang2020tablesurvey,zhu-etal-2021-tat,pasupat:15,Abbas2016WikiQAA,sun2016table,chen2021kace,chen2020hybridqa,lin2020bridging,zayats2021representations, oguz2020unified} and table-to-text generation \citep{zhang2020tablesum,parikh2020totto, radev2020dart,yoran2021turning,chen2020open}.

Similar to our data setting, some recent papers have also proposed ideas for representing Wikipedia relational tables, some such papers are TAPAS \citep{herzig-etal-2020-tapas}, StrucBERT \citep{trabelsi2022structbert}, Table2vec \citep{zhang2020table2vec}, TaBERT \citep{yin-etal-2020-tabert}, TABBIE \citep{iida-etal-2021-tabbie},TabStruc \citep{zhang-etal-2020-table}, TabGCN \citep{pramanick2021joint}, RCI \citep{glass-etal-2021-capturing}, TURL \citep{10.1145/3542700.3542709} and TableFormer \citep{yang2022tableformer}. Some papers such as \citep[ and others]{yu2018spider,yu2020grappa,eisenschlos:20, neeraja-etal-2021-infotabskg, Mller2021TAPASAS, somepalli2021saint} study the improvement of tabular inference by pre-training.

\paragraph{Tabular Datasets.} Synthetic creation of dataset has long been explored \citep[ and others]{Rozen2019DiversifyYD,Mller2021TAPASAS,Kaushik2020Learning,Xiong2020Pretrained}. For tabular NLI in particular, the datasets can be categorized into 1) Manually created datasets ~\citep{gupta-etal-2020-infotabs} with manually creates both hypothesis and premise, \citep{chen2019tabfact} manually creates the hypothesis while premise is automatically generated 2) Synthetically created semi-automatically generated datasets which completely automate data generation requires manual designing table-dependent context-free grammar (CFG) \cite{eisenschlos:20}, or require logical forms to be annotated \cite{Mller2021TAPASAS, chen-etal-2020-logical,chen-etal-2020-logic2text}. Several works such as \citet{poliak2018hypothesis,niven-kao-2019-probing, gururangan2018annotation,glockner-etal-2018-breaking,naik2018stress, Wallace2019UniversalAT} have shown that models exploit spurious patterns in data. Similar to \citet{nie2019analyzing, Zellers2018SWAGAL,gupta-etal-2020-infotabs} authors investigate impacts of artifacts in dataset by creating adversarial test sets. However, semi-automatic systems requiring a CFG or logical forms contains reasoning which is often limited to certain types. Creating sentences that contain other reasonings (like lexical reasoning, knowledge, and common sense reasoning) is challenging using CFG and logical forms. Our paper requires subject matter experts to create entity specific templates for each category which leads to creating sentences with multiple reasonings as well as complex reasonings.

\section{Discussion}
\label{sec:more_discussion}
\paragraph{Why Counterfactual Table Generation?} Tabular dataset is inherently semi-structured. Therefore, each category table has a specific set of keys. This enables us to create key-specific templates based on the entity-types of keys \citep{neeraja-etal-2021-infotabskg}, which could be applied to millions of tables of a given category. Furthermore, as explained in \S\ref{sec:autotnlidataset}, the templates also generalize across keys with similar value types across categories. All this is only possible due to the semi-structured nature of tabular data. Using counterfactual tables equips the model with more linguistically comparable. But oppositely labeled data to learn from, guaranteeing that the model can learn beyond the superficial textual artifacts and so becomes more resilient as shown by \citep{rajagopal2022counterfactual,Kaushik2020Learning}. As a result, when counterfactual data is included in the \ourdatasetName, we observe performance improvement throughout all experimental settings. This learning is further verified by the findings for better gains in \alphaTwo, which comprises instances of linguistically comparable but oppositely labeled data instances.

\paragraph{Why Semi-Automatic Approach?} By examining the two diametrically opposed frameworks, namely a Human and an Automatic Annotation Framework, we may see many issues with both. Manually created data is prohibitively expensive and demands much human effort, limiting the ability to develop large-scale databases. Additionally, humans have a propensity to establish artificial patterns when manually creating a dataset, such as not giving all keys the same importance (explained in \S\ref{sec:autotnlidataset}). 
While autonomous data generation is computationally efficient, it has many limitations. e.g., the inability to generate linguistically complex sentences and the difficulty of incorporating reasoning into the dataset. Because most deep learning models perform better with more data, producing large-scale datasets at a reasonable cost is critical while retaining data quality. With this in mind, we presented a "semi-automatic" architecture with the following benefits: \begin{inparaenum}[(a.)] \item It simplifies the creation of large-scale datasets. Using only 660 templates, we can generate 1,478,662 premise-hypothesis pairings from around 10,182 tables. \item The framework may be reused with additional tabular data as long as the structure is preserved. \item It enables the creation of linguistically and lexically diverse datasets. \item As shown in \S\ref{sec:autotnlidataset}, hypothesis bias can be minimized by establishing an adequate number of diverse templates for all keys of each category. \item The premises have been paraphrased in three ways to bring the required lexical diversity.\end{inparaenum}

\section{Conclusion}
\label{sec:conclusion}

We introduced a semi-automatic framework for generating data from tabular data. Using a template-based approach, we generate \ourdatasetName. We utilized \ourdatasetName for both TNLI evaluation and data augmentation. Our experiments demonstrate the effectiveness of \ourdatasetName and, by implication, our framework, especially for adversarial settings. For the future work, we aim to involve the creation of additional lexically varied and robust datasets and investigate whether the addition of neutrals could improve these datasets.

\section{Limitations}
\label{sec:limitation}

This work has focused on entity tables, where the tabular structure and knowledge patterns are straightforward. 
Nevertheless, our templates technique does not generate maybe true/maybe false statements, i.e., neutral statements, as they need enhanced common sense (e.g., subjective usage) and unmentioned entity knowledge, i.e., information beyond the premise table. 
It is unknown how to generate good templates automatically, such as using neural generation methods rather than leveraging expert domain knowledge. 
Also, how these manually curated templates work when applied with more complicated tables like nested and hierarchical tables is under-explored. 
Theoretically, we can generate an infinite number of premise-hypothesis pairs, but the marginal utility might hurt the notion. 
Additionally, the zero-shot capabilities for out-of-domain tables are limited by the presumption that tables in similar categories resemble keys.

\section*{Acknowledgement}
We thank members of the Utah NLP group for their valuable insights and suggestions at various stages of the project; and reviewers their helpful comments. We thank Antara Bahursettiwar for her valuable feedback. Additionally, we appreciate the inputs provided by Vivek Srikumar and Ellen Riloff. Vivek Gupta acknowledges support from Bloomberg’s Data Science Ph.D. Fellowship.

\bibliography{anthology,custom}
\bibliographystyle{acl_natbib}

\appendix

\label{sec:appendix}
\section{Cross-Category Analysis}
We analyze how the semi-automatic data created performs across categories, i.e., training on one category and evaluating on the rest. This gave an idea of how training on data from one category generalizes over the rest. In Table \ref{Appendix1}, we have shown the accuracy when our model is trained on the categories written in rows and evaluated on the categories given in the columns.
\begin{table*}[!h]
\small
\centering
\begin{tabular}{lccccccccccc}
\hline
\textbf{Category} & \textbf{City}  & \textbf{Album} & \textbf{Person} & \textbf{Movie} & \textbf{Book} & \textbf{F\&D} & \textbf{Org} & \textbf{Paint} & \textbf{Fest} & \textbf{S\&E} & \textbf{Univ} \\
\hline
\textbf{City} & 88.64 & \textcolor{red}{51.85} & 70.34 & 77.29 & 77 & 68.48 & 75.05 & 70.73 & 75.98 & 66.75 & 77.43 \\
\textbf{Album} & 52.92 & 79.35 & \textcolor{darkgreen}{65.2} & \textcolor{darkgreen}{60.28} & \textcolor{darkgreen}{57.38} & \textcolor{darkgreen}{65.75} & \textcolor{darkgreen}{59.16} & \textcolor{darkgreen}{53.48} & \textcolor{darkgreen}{58.8} & \textcolor{darkgreen}{55.75} & \textcolor{red}{52.9} \\
\textbf{Person} & 75.57 & \textcolor{red}{57.57} & 94.58 & 89.72 & 91.02 & 81.99 & 83.86 & 80.52 & 86.01 & 69.58 & 81.25 \\
\textbf{Movie} & 76.49 & \textcolor{red}{56.97} & 85.41 & 98.26 & 87.01 & 82.11 & 84.65 & 71.29 & 84.79 & 69.34 & 81.01 \\
\textbf{Book} & 54.03 & \textcolor{red}{53.37} & 76 & 77.69 & 97.84 & 78.68 & 76.81 & 73.51 & 64.94 & 71.62 & 53.76 \\
\textbf{F\&D} & 61.79 & \textcolor{red}{56.72} & 80.67 & 83.24 & 87.55 & 95.82 & 80.46 & 76.49 & 74.61 & 68.71 & 58.03 \\
\textbf{Org} & 74.73 & \textcolor{red}{55.89} & 83.67 & 88.26 & 85.08 & 80.64 & 96.36 & 70.72 & 83.85 & 68.84 & 81.22 \\
\textbf{Paint} & 54.24 & \textcolor{violet}{50.45} & 65.71 & 70.39 & 73.41 & 68.3 & 64.52 & 99 & 59.58 & 61.52 & 54.44 \\
\textbf{Fest} & 73.4 & \textcolor{red}{52.46} & 82.65 & 87.77 & 81.98 & 78.23 & 80.02 & 72.27 & 88.49 & 64.83 & 77.3 \\
\textbf{S\&E} & \textcolor{violet}{51.52} & 53.53 & 69.15 & 73.52 & 85.75 & 72.49 & 70.23 & 76.24 & 61.86 & 95.39 & \textcolor{darkgreen}{52.17} \\
\textbf{Univ} & 76.06 & \textcolor{red}{51.16} & 78.67 & 85.03 & 76.26 & 76.99 & 78.46 & 68.18 & 79.77 & 69.91 & 91.9 \\
\hline
\end{tabular}
\caption{\label{Appendix1} Cross-category analysis of our data. \textcolor{red}{red} - shows the least accuracy when trained on a category and evaluated on another. \textcolor{darkgreen}{green} - the least accuracy obtained when tested on a category and trained on the others. \textcolor{violet}{violet} - intersection of the two cases above (\textbf{F\&D}- Food \& Drinks, \textbf{S\&E} - Sports \& Events)}
\end{table*}

\textbf{Analysis:} Here we observed that except some categories such as \emph{Sports \& Events}, \emph{Album} and \emph{City} the cross category accuracy is pretty high among the rest. \emph{Album} seems to be quite a hard category with all categories giving a low cross-category accuracy when evaluated on it. \emph{City} gave a challenging test set when trained on \emph{Sport \& Events}. \emph{University} is the toughest test set for \emph{Album}. When used as a test-set, \emph{City} gave the least accuracy against \emph{Sports \& Events}, \emph{Album} gives the least accuracy against \emph{Paint}, \emph{University} gave the least accuracy against \emph{Sports \& Events} and for the rest \emph{Album} gave the least accuracy.

\section{Cross-Entity Analysis}
We analyze how the semi-automatic data created performs across entities, i.e., training on one entity and evaluating on the rest. This gave an idea of how training on data from one category generalizes over the rest. In Table \ref{Appendix2}, we have shown the accuracy when our model is trained on the entity written in rows and evaluated on the entities given in the columns.
\begin{table*}[!h]
\small
\centering
\begin{tabular}{lccccccccccc}
\hline
\textbf{Entity} & \textbf{Person}  & \textbf{P\&T} & \textbf{Skill} & \textbf{Org} & \textbf{Quantity} & \textbf{D\&T} & \textbf{Location} & \textbf{Event} & \textbf{URL} & \textbf{Product} & \textbf{Other} \\
\hline
\textbf{Person} & 98.44 & 81.24 & 85.56 & 84.5 & 68.83 & \textcolor{red}{61.59} & 84.77 & 84.97 & 76.14 & 86.1 & 78.74 \\
\textbf{P\&T} & 70.45 & 98.33 & 68.77 & 67.84 & 55.58 & \textcolor{red}{55.42} & 64.77 & 78.26 & \textcolor{darkgreen}{58.94} & 67.17 & 71.1 \\
\textbf{Skill} & 79.44 & 88.01 & 93.44 & 79.92 & \textcolor{red}{53.76} & 57.65 & 78.48 & 89.18 & 73.04 & 82.29 & 73.13 \\
\textbf{Org} & 92.36 & 87.33 & 86.58 & 95.62 & 63.56 & \textcolor{red}{58.03} & 87.19 & 87.12 & 84.09 & 86.9 & 81.29 \\
\textbf{Quantity} & 82.12 & \textcolor{red}{61.93} & 67.27 & 71.41 & 91.36 & 63.22 & 78.13 & 77 & 78.97 & 70.71 & 70.62 \\
\textbf{D\&T} & 77.27 & 65.01 & \textcolor{red}{60.18} & 74.98 & 64.39 & 85.87 & 77.28 & 71.19 & 88.93 & 64.78 & 70.02 \\
\textbf{Location} & 88.32 & 76.32 & 86.3 & 83.18 & 68.89 & \textcolor{red}{62.31} & 94.43 & 81.57 & 83.69 & 79.98 & 75.75 \\
\textbf{Event} & 86.01 & 76.66 & 79.52 & 79.8 & 66.14 & \textcolor{red}{57.17} & 79.75 & 97.09 & 79.05 & 77.92 & 75.6 \\
\textbf{URL} & \textcolor{darkgreen}{61} & \textcolor{darkgreen}{56.27} & \textcolor{darkgreen}{58.42} & \textcolor{darkgreen}{60.88} & \textcolor{violet}{51.61} & \textcolor{darkgreen}{55.02} & \textcolor{darkgreen}{62.68} & \textcolor{darkgreen}{60.56} & 95.25 & \textcolor{darkgreen}{56.07} & \textcolor{darkgreen}{55.09} \\
\textbf{Product} & 88.82 & 84.03 & 87.59 & 85.5 & 67.24 & \textcolor{red}{62.11} & 87.02 & 89.83 & 77.77 & 98.99 & 77.37 \\
\textbf{Other} & 83.39 & 84.98 & 80.82 & 78.24 & 62.44 & \textcolor{red}{58.29} & 76.97 & 86.74 & 69.98 & 82.78 & 93.88 \\
\hline
\end{tabular}
\caption{\label{Appendix2} Cross-entity analysis of our data. \textcolor{red}{red} - shows the least accuracy when trained on a entity and tested on another. \textcolor{darkgreen}{green} - the least accuracy obtained when tested on an entity and trained on the others. \textcolor{violet}{violet} - intersection of the two cases above (\textbf{P\&T}- Person Type, \textbf{D\&T} - Date \& Time)}
\vspace{-1.5em}
\end{table*}

\textbf{Analysis:} Here we observed that \emph{Date \& Time} is quite a tough test-set for most entities. \emph{Quantity} is a tough test-set for \emph{Skill} and \emph{URL}. For \emph{Skill} and \emph{Person Type} are tough test-sets for \emph{Location} and \emph{Quantity} respectively. When used as a test-set, \emph{URL} gave the lowest accuracy against \emph{Person Type}, \emph{Quantity} gave the lowest accuracy against \emph{URL} and for the rest the \emph{URL} gave the least accuracy.

\section{First Stage Performance with Limited Supervision}
The first stage classifier is used to classify \neutral vs. \nonneutral. In Table \ref{tab:limited_setting_first_stage} we have shown the accuracy for the first stage of the 2-stage classifier in the limited supervision setting with and w/o MNLI augmentation.

\begin{table}[!h]
\small
\centering
\begin{tabular}{ccc|ccc}
\hline
\textbf{Tr(\%)} & \textbf{No Aug} & \textbf{MNLI} &\textbf{Tr(\%)} & \textbf{No Aug} & \textbf{MNLI}\\
\hline
\multicolumn{3}{c}{\bf Development set} & \multicolumn{3}{|c}{\bf \alphaTwo set}\\
\hline
0 & \bf 63.11 & 59.28 & 0 & 64.61 & \bf 65.33 \\
5 & 75.75 & \bf 81.42 & 5 & 78.17 & \bf 84.03 \\
10 & 76.86 & \bf 83.08 & 10 & 79.86 & \bf 85.53 \\
15 & 78.92 & \bf 83.03 & 15 & 81 & \bf 85.72 \\
20 & 78.83 & \bf 82.83 & 20 & 81.58 & \bf 85.89 \\
25 & 78.92 & \bf 83.47 & 25 & 81.81 & \bf 85.89 \\
\hline
\multicolumn{3}{c}{\bf \alphaOne set} & \multicolumn{3}{|c}{\bf \alphaThree set}\\
\hline
0 & \bf 62.28 & 58.5 & 0 & \bf 62.72 & 56.06 \\
5 & 76.94 & \bf 81.86 & 5  & 69.42 & \bf 72.78 \\
10 & 77.11 & \bf 82.67 & 10  & 69.17 & \bf 72.97 \\
15 & 79.22 & \bf 82.53 & 15  & 70.22 & \bf 73.44 \\
20 & 78.53 & \bf 82.56  & 20  & 67.86 & \bf 73.56 \\
25 & 78.92 & \bf 82.78 & 25  & 68.81 & \bf 74.03 \\
\hline
\end{tabular}
\caption{First stage performance (accuracy) of RoBERTa$_{\text{BASE}}$ (i.e. \neutral or \nonneutral) classifier with various data augmentation for limited supervision setting i.e. varying percentage of \datasetName training data. The average standard deviation across 3 runs is 1.197 with range varying from 0$\%$ to 3.14$\%$. \textbf{bold} same as Table \ref{tab:RQ2}.}
\label{tab:limited_setting_first_stage}
\end{table}

\section{Effects of augmentation with \ourdatasetName  in limited supervision}

Since \ourdatasetName only contains \entail and \contradict labels, to check how pretraining with \ourdatasetName affects the results in the limited supervision setting we had to use the 2-stage classifier where \begin{inparaenum}[(a.)] \item No Augmentation in first stage i.e. Table \ref{tab:limited_setting_both_stages}. \item Augmentation with MNLI in first stage i.e. Table \ref{tab:limited_setting_both_stages_with_MNLI}. \end{inparaenum} 

\begin{table}[!h]
\small
\centering
\begin{tabular}{c|ccccc}
\hline
   &   &  & & \textbf{MNLI} & \textbf{MNLI} \\
\textbf{Tr(\%)} &\textbf{No} & \textbf{Orig} & \textbf{Orig} & \textbf{+Orig} & \textbf{+Orig} \\
 &\textbf{Aug} & & \textbf{+Count} & & \textbf{+Count}\\
\hline
& \multicolumn{5}{c}{\bf Development set}\\
\hline
0 & 33.06 & 38.94 & 34.5 & \bf 39.56 & \bf 39.56 \\
5 & 55.64 & 57.44 & \bf 59.11 & 58.31 & 58.42 \\
15 & 61.83 & 62.22 & \bf 63.44 & 63.42 & 63.28 \\
25 & 64.5 & 64.83 & 64.97 & 65.47 & \bf 66.11 \\
\hline
& \multicolumn{5}{c}{\bf \alphaOne set}\\
\hline
0 & 33 & 38.06 & 34.33 & \bf 40.28 & \bf 40.28 \\
5 & 56.64 & 58.75 & 58.89 & 58.67 & \bf 59.03 \\
15 & 61.81 & 62.94 & 63.03 & \bf 64.33 & 63.89 \\
25 & 64.36 & 64.39 & 64.28 & 64.89 & \bf 65.69 \\
\hline
& \multicolumn{5}{c}{\bf \alphaTwo set}\\
\hline
0 & 33.17 & 38.83 & 39.5 & \bf 40.5 & \bf 40.5 \\
5 & 43.69 & 47.64 & 49.64 & 49.86 & \bf 51.44 \\
15 & 50.39 & 53.69 & 53.72 & 54.75 & \bf 54.94 \\
25 & 54.69 & 56 & 56.03 & 57.06 & \bf 57.94 \\
\hline
& \multicolumn{5}{c}{\bf \alphaThree set}\\
\hline
0 & 32.83 & 38.39 & 36.61 & \bf 41.61 & \bf 41.61 \\
5  & 47.42 & 48.31 & 51.03 & 50.94 & \bf 52.31 \\
15  & 50.47 & 52.28 & 53.44 & \bf 53.86 & 53.19 \\
25  & 52.33 & 51.81 & 52.06 & \bf 54.25 & 53.83 \\
\hline
\end{tabular}
\caption{Both stage performance (accuracy) of RoBERTa$_{\text{BASE}}$ (i.e. \entail, \contradict or \neutral) classifier with various data augmentation for limited supervision setting i.e. varying percentage of \datasetName training data w/o MNLI pretraining for first stage. The average standard deviation across 3 runs is 0.98 with range varying from 0$\%$ to 4$\%$. \textbf{bold} same as Table \ref{tab:RQ2}.}
\label{tab:limited_setting_both_stages}
\vspace{-1.5em}
\end{table}

\begin{table}[!h]
\small
\centering
\begin{tabular}{c|ccccc}
\hline
   &   &  & & \textbf{MNLI} & \textbf{MNLI} \\
\textbf{Tr(\%)} &\textbf{No} & \textbf{Orig} & \textbf{Orig} & \textbf{+Orig} & \textbf{+Orig} \\
 &\textbf{Aug} & & \textbf{+Count} & & \textbf{+Count}\\
\hline
& \multicolumn{5}{c}{\bf Development set}\\
\hline
0 & 43.33 & 47 & 45.44 & \bf 47.94 & 47.72 \\
5 & 61.11 & 63.25 & \bf 64.81 & 64.19 & 64.36 \\
15 & 65.33 & 65.67 & 66.58 & 66.86 & \bf 66.89 \\
25 & 68.08 & 68.25 & 68.19 & 69.06 & \bf 69.56 \\
\hline
& \multicolumn{5}{c}{\bf \alphaOne set}\\
\hline
0 & 42.06 & 47.94 & 45.78 & \bf 48.17 & 48.11 \\
5 & 61.83 & 64.06 & 64.08 & 64 & \bf 64.36 \\
15 & 64.39 & 65.72 & 65.58 & \bf 66.97 & 66.61 \\
25 & 67.69 & 68.03 & 67.61 & 68.17 & \bf 69.03 \\
\hline
& \multicolumn{5}{c}{\bf \alphaTwo set}\\
\hline
0 & 46.72 & 51.61 & 50.5 & 51.5 & \bf 52.06 \\
5 & 49.69 & 53.78 & 56.17 & 56.39 & \bf 57.67 \\
15 & 54.47 & 57.83 & 57.81 & 58.97 & \bf 59.14 \\
25 & 57.31 & 59.17 & 59.22 & 60.11 & \bf 61.08 \\
\hline
& \multicolumn{5}{c}{\bf \alphaThree set}\\
\hline
0 & 39.72 & 43.33 & 42.33 & \bf 44.72 & 44.17 \\
5  & 50.64 & 51.67 & 54.14 & 54.56 & \bf 56.22 \\
15  & 53.72 & 55.58 & 56.67 & \bf 57.19 & 56.75 \\
25  & 56 & 55.39 & 55.89 & \bf 58.69 & 57.92 \\
\hline
\end{tabular}
\caption{Both stage performance (accuracy) of RoBERTa$_{\text{BASE}}$ (i.e. \entail, \contradict or \neutral) classifier with various data augmentation for limited supervision setting i.e. varying percentage of \datasetName training data with MNLI pretraining for first stage. The average standard deviation across 3 runs is 1.89 with range varying from 0$\%$ to 5.23$\%$. \textbf{bold} same as Table \ref{tab:RQ2}.}
\label{tab:limited_setting_both_stages_with_MNLI}
\vspace{-1.0em}
\end{table}

\textbf{Analysis:} As we can see in both Table \ref{tab:limited_setting_both_stages} and Table \ref{tab:limited_setting_both_stages_with_MNLI} that the best is obtained by similar models in either case, with the only difference being that augmenting the first stage with MNLI helps improve the accuracy across all cases.

\section{Automatic Data Generation}
\label{sec:appendi_auto-gen}
Using GPT-J-6B, we generate 9–11 sentences per category. In total, we generated 110 sentences for 11 categories. We then classified each sentence into one of the following five classes: \begin{inparaenum}[(a.)] \item Correct - Both sentence and labels are correct. \item Factual error - Sentence is meaningful, but the label assigned to it is wrong. \item Overfit error - The same sentence as seen previously is generated without any lexical changes. \item Hallucination error - When knowledge from outside the tables provided is used to make a sentence. \item Repetition error - The same sentence is generated several times.\end{inparaenum}
\newline
\begin{figure}[!h]
    \centering
    \includegraphics[width=0.50\textwidth]{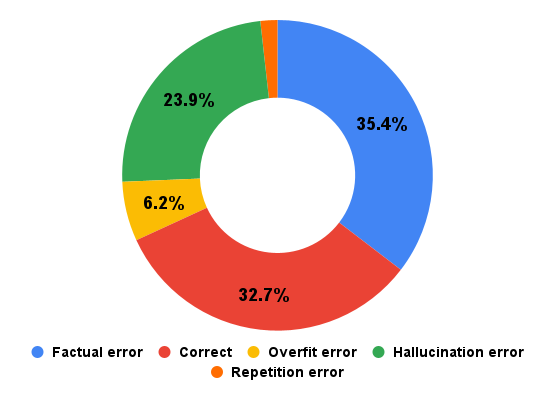}
    \caption{Percentage chart for Automatic data generation correct and error labels.}
    \label{fig:auto_gen}
\end{figure}
\textbf{Analysis}: As observed in Figure \ref{fig:auto_gen}, out of all the 110 automatically generated hypothesis only 32.7\% were \emph{Correct} i.e. sentences were meaningful and the labels assigned to them are correct. Among the rest, about 52\% had \emph{Factual} errors in them and around 35\% were \emph{Hallucination} errors. This further demonstrates that a semi-automatic approach, such as ours, is preferable, as fully automated generating techniques are not reliable.
\\

\section{ALBERT Performance}
\label{sec:appendixALBERT}
We perform a similar analysis on ALBERT$_{\text{BASE}}$ as we have done for RoBERTa$_{\text{BASE}}$ to see if our data benefits there too. To see how robust \ourdatasetName is when improving performance in the Augmentation setting, we perform the same experiments as RQ2a in Section \ref{sec:Data_aug}. We also explore some experiments from RQ1b in Section \ref{sec:RQ1} which are shown in Table \ref{tab:albert_RQ1}.
\newline
\begin{table}[!h]
\small
\centering
\label{fig:Ablation-stage1} 
\begin{tabular}{l|cc}
\toprule
\textbf{Test-split} &\bf No-Augmentation  & \textbf{MNLI} \\
\midrule
dev & 79.11 & \bf 85.22 \\
\alphaOne & 78.61 & \bf 82.83 \\
\alphaTwo & 80.89 & \bf 85.22 \\
\alphaThree & 67.78 & \bf 73.94 \\
\bottomrule
\end{tabular}
\vspace{-0.5em}
\caption{Performance (accuracy) of stage one ALBERT$_{\text{BASE}}$ (i.e. \neutral vs \nonneutral) across several data augmentation settings. Here, No-Augmentation means \datasetName, and MNLI means MNLI + \datasetName. \textbf{bold} same as Table \ref{tab:RQ2}.}
\label{tab:albert_AblationStage1}
\vspace{-0.5em}
\end{table}

\begin{table}[!h]
\vspace{-0.5em}
\small
\centering
\begin{tabular}{l|ccccc}
\hline
& \multicolumn{5}{|c}{\textbf{Stage 2: Entail  vs Contradict}} \\
\hline
 &   &  & & \textbf{MNLI} & \textbf{MNLI} \\
\textbf{Split} &\textbf{No} & \textbf{Orig} & \textbf{Orig} & \textbf{+Orig} & \textbf{+Orig} \\
 &\textbf{Aug} & & \textbf{+Count} & & \textbf{+Count}\\
\hline
& \multicolumn{5}{c}{\bf Stage 1: \datasetName} \\
\hline
dev & 60.78 & 61.72 & 62.83 & \bf 64.83 & 63.89 \\
\alphaOne & 60.89 & 61.33 & 62.78 & \bf 63.22 & 63.11 \\
\alphaTwo & 49.83 & 53.06 & 51.67 & 55.67 & \bf 56.5 \\
\alphaThree & 49.39 & 50.11 & 51.72 & \bf 52.94 & 51.72 \\
\hline
& \multicolumn{5}{c}{\bf Stage 1: MNLI +\datasetName}\\
\hline
dev & 66.28 & 67.44 & 68.22 & \bf 70.67 & 69.61 \\
\alphaOne & 65.72 & 66.06 & 67.28 & 67.44 & \bf 67.5 \\
\alphaTwo & 54 & 56.72 & 55.83 & 60.11 & \bf 60.89 \\
\alphaThree & 53.33 & 55.11 & 56.11 & \bf 57.39 & 56.94 \\
\hline
\end{tabular}
\vspace{-0.5em}
\caption{Accuracy of combine stage I i.e. \neutral vs \nonneutral and stage II i.e. \entail vs \contradict classifiers (ALBERT$_{\text{BASE}}$) across several data augmentation settings. Here, for stage one we also explore pre-fine tuning on MNLI data. \textbf{bold} - represents max across columns i.e. the best augmentation setting.}
\label{tab:albert_RQ2}
\vspace{-1.0em}
\end{table}

\begin{table}[!h]

\small
\centering
\begin{tabular}{l|ccccc}
\hline
 &   &  & & \textbf{MNLI} & \textbf{MNLI} \\
\textbf{Split} &\textbf{No} & \textbf{Orig} & \textbf{Orig} & \textbf{+Orig} & \textbf{+Orig} \\
 &\textbf{Aug} & & \textbf{+Count} & & \textbf{+Count}\\
\hline
dev & 68.92 & 71.25 & 72.33 & \bf 76 & 74.5 \\
\alphaOne & 69.42 & 70.92 & 72.92 & \bf 73.92 & 73.25 \\
\alphaTwo & 47.58 & 52.75 & 50.83 & 58.17 & \bf 58.67 \\
\alphaThree & 61 & 64.17 & 66.33 & \bf 68.33 & 68.08 \\
\hline
\end{tabular}
\vspace{-0.5em}
\caption{Accuracy of stage II i.e. \entail vs \contradict classifiers (ALBERT$_{\text{BASE}}$) across several data augmentation settings. \textbf{bold} same as Table \ref{tab:RQ2}. }
\label{tab:albert_AblationStage2}
\vspace{-1.5em}
\end{table}
\noindent \textbf{Analysis:} As we can see in Table \ref{tab:albert_AblationStage1} to Table \ref{tab:albert_AblationStage2}, the trends are very similar to what we have seen in main paper Section \ref{sec:Data_aug} for full supervision setting. Thus our approach of semi-automatic generation is generalizable across similar models.

\begin{table*}[!h]
\small
\centering
\begin{tabular}{l|cccccc}
\toprule
 \textbf{Augmentation Strategy} & \textbf{Cat-Ran} & \textbf{Cross-Cat} & \textbf{Key} & \textbf{No-Para} & \textbf{Cross-Para} & \textbf{Entity}\\
\midrule
Random & 50.00 & 50.00 & 50.00 & 50.00 & 50.00 & 50.00 \\
\ourdatasetName & 77.16 & 69.73 & 81.91 & 86.22 & \bf 87.45 & 72.75 \\
MNLI +\ourdatasetName & \bf 80.28 & \bf 76.24 & \bf 83.1 & \bf 88.73 & 87.44 & \bf 74.53 \\
\bottomrule
\end{tabular}
\vspace{-1.0em}
\caption{Performance (accuracy) on \ourdatasetName with ALBERT$_{\text{BASE}}$ model across several evaluation splits with fine-tuning on \ourdatasetName. \textbf{bold} - represents max across rows i.e. best train/augmentation setting.}
\label{tab:albert_RQ1}
\vspace{-1.0em}
\end{table*}







\section{Performance Across Different Reasoning Types in \datasetName}
\label{sec:appendix_reasoning}
We take the 160 pairs from development and \alphaThree test sets each, from \datasetName, that have been categorised into 14 reasoning types to assess the impact of pre-training on various reasoning types, namely \begin{inparaenum}[(a)] \item numerical reasoning, \item co-reference, \item multi-row reasoning, \item knowledge and common sense, \item simple lookup, \item negation, \item lexical reasoning, \item entity type, \item named entities, \item temporal reasoning, \item subjective/out-of-table, \item quantification, \item syntactic alternations, and \item ellipsis. \end{inparaenum} The frequency of \entail and \contradict pairs being correctly classified is shown in Table \ref{tab:reasoning_type_infotabs_entail} and Table \ref{tab:reasoning_type_infotabs_contradict} respectively.

\textbf{Analysis:} In Table \ref{tab:reasoning_type_infotabs_entail} we observe that 9 out of 14 times in development and 12 out of 14 times in \alphaThree-test sets MNLI + Orig + Count perform best. In Table \ref{tab:reasoning_type_infotabs_contradict} we observe that 10 out of 14 times in development set Orig + Count perform best.  

\begin{table*}[!h]
\small
\centering
\begin{tabular}{c|c|ccc|c|ccc}
\hline
& & & & \textbf{MNLI} & & & & \textbf{MNLI}  \\
& \textbf{Human} & \textbf{No} & \textbf{Orig} & \textbf{+Orig} & \textbf{Human} & \textbf{No} & \textbf{Orig} & \textbf{+Orig} \\
& & \textbf{Aug} & \textbf{+Count} & \textbf{+Count} & & \textbf{Aug} & \textbf{+Count} & \textbf{+Count} \\
\hline
&\multicolumn{4}{c}{\bf Development set} &\multicolumn{4}{|c}{\bf \alphaThree set}\\
\hline
numerical & 11 & 6 & \bf 8 & \bf 8 & 14 & 1 & 3 & \bf 5 \\
co-reference & 8 & \bf 4 & \bf 4 & 3 & 5 & 2 & 2 & \bf 3 \\
multi-row & 20 & \bf 13 & 11 & \bf 13 & 15 & 6 & \bf 8 & \bf 8 \\
KCS & 31 & 18 & \bf 21 &\bf 21 & 11 & 6 & \bf 9 & 8 \\
temporal & 19 & 10 & 15 & \bf 16 & 10 & 6 & 7 & \bf 8 \\
syntactic-alt & 0 & 0 & 0 & 0 & 2 & 1 & 1 & \bf 2 \\
simple-lookup & 3 & \bf 3 & \bf 3 & \bf 3 & 8 & \bf 8 & 7 & \bf 8 \\
entity-type & 6 & 4 & \bf 5 & 4 & 8 & 3 & \bf 6 & \bf 6 \\
ellipsis & 0 & 0 & 0 & 0 & 1 & \bf 0 & \bf 0 & \bf 0 \\
subjective-oot & 6 & 3 & \bf 4 & \bf 4 & 2 & \bf 1 & \bf 1 & \bf 1 \\
name-id & 2 & \bf 1 & \bf 1 & \bf 1 & 1 & \bf 1 & \bf 1 & \bf 1 \\
lexical & 5 & \bf 3 & \bf 3 & \bf 3 & 3 & 2 & \bf 3 & \bf 3 \\
quantification & 4 & 1 & \bf 3 & \bf 3 & 2 & \bf 2 & \bf 2 & \bf 2 \\
negation & 0 & 0 & 0 & 0 & 0 & 0 & 0 & 0 \\
\hline
\end{tabular}
\vspace{-1.0em}
\caption{Frequency of labels assigned as \entail in each reasoning type across 3 settings and Gold labels for \datasetName. \textbf{bold} - represents max across rows i.e. best train/augmentation setting. }
\label{tab:reasoning_type_infotabs_entail}
\vspace{-0.7em}
\end{table*}

\begin{table*}[!h]
\small
\centering
\begin{tabular}{c|c|ccc|c|ccc}
\hline
& & & & \textbf{MNLI} & & & & \textbf{MNLI}  \\
& \textbf{Human} & \textbf{No} & \textbf{Orig} & \textbf{+Orig} & \textbf{Human} & \textbf{No} & \textbf{Orig} & \textbf{+Orig} \\
& & \textbf{Aug} & \textbf{+Count} & \textbf{+Count} & & \textbf{Aug} & \textbf{+Count} & \textbf{+Count} \\
\hline
&\multicolumn{4}{c}{\bf Development set} &\multicolumn{4}{|c}{\bf \alphaThree set}\\
\hline
numerical & 7 & \bf 5 & \bf 5 & \bf 5 & 14 & \bf 12 & 10 & 7 \\
co-reference & 13 & 8 & \bf 10 & 8 & 8 & \bf 6 & 5 & 4 \\
multi-row & 17 & \bf 12 & \bf 12 & \bf 12 & 12 & \bf 10 & 8 & 8 \\
KCS & 24 & 15 & \bf 17 & 16 & 17 & \bf 12 & \bf 12 & \bf 12 \\
temporal & 25 & 15 & \bf 18 & 15 & 16 & \bf 14 & 11 & 12 \\
syntactic-alt & 0 & 0 & 0 & 0 & 0 & 0 & 0 & 0 \\
simple-lookup & 1 & \bf 0 & \bf 0 & \bf 0 & 2 & \bf 2 & \bf 2 & \bf 2 \\
entity-type & 6 & 3 & \bf 4 & \bf 4 & 9 & \bf 4 & 3 & 1 \\
ellipsis & 0 & 0 & 0 & 0 & 0 & 0 & 0 & 0 \\
subjective-oot & 6 & 2 & \bf 3 & 2 & 9 & \bf 5 & 4 & 3 \\
name-identity & 1 & \bf 1 & \bf 1 & \bf 1 & 0 & 0 & 0 & 0 \\
lexical & 4 & \bf 4 & 3 & \bf 4 & 8 & \bf 5 & 4 & 3 \\
quantification & 6 & 3 & \bf 4 & \bf 4 & 4 & \bf 2 & 1 & \bf 2 \\
negation & 6 & \bf 6 & \bf 6 & \bf 6 & 4 & \bf 3 & \bf 3 & 2 \\
\hline
\end{tabular}
\vspace{-1.0em}
\caption{Frequency of labels assigned as \contradict in each reasoning type across 3 settings and Gold labels for \datasetName. \textbf{bold} - represents max across rows i.e. best train/augmentation setting. }
\label{tab:reasoning_type_infotabs_contradict}
\vspace{-1.0em}
\end{table*}

\section{Reasoning for \ourdatasetName}
\label{sec:appendix_reasoning_autotnli}

\begin{figure}[!htbp]
    \centering
    \includegraphics[width=0.45\textwidth]{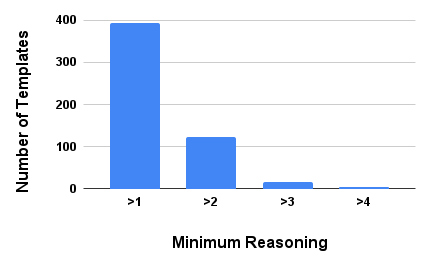}
    \vspace{-1.5em}
    \caption{Cumulative frequency of templates across reasoning types in \ourdatasetName.}
    \label{fig:appendix_reasoning_stat_4}
\end{figure}

Our annotators classified all the distinct\footnote{     templates for Provost and President are very similar so we don't consider them to be separate templates} templates from \ourdatasetName into 14 reasoning types present in \datasetName. Table \ref{tab:appendix_reasoning_stat_2} shows the individual reasoning type distribution across each category. The distribution statistics of reasoning types across each category is shown in Table \ref{tab:appendix_reasoning_stat_1}. Table \ref{tab:appendix_reasoning_stat_3} shows that summary statistics across various reasoning types. Figure \ref{fig:appendix_reasoning_stat_4} gives distribution of extend of multiple reasoning in each individual examples.

\begin{figure*}[h]
    \centering
    \begin{subfigure}[t]{0.49\textwidth}
        \centering
        \includegraphics[width=\textwidth]{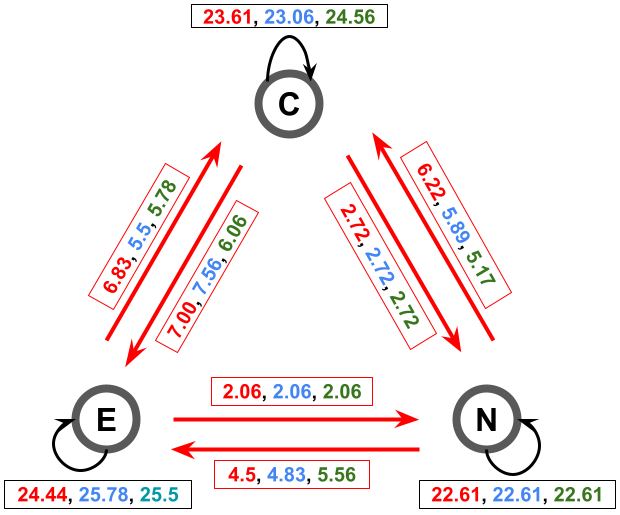}
        \caption{development}
    \end{subfigure}
    \hfill
    \begin{subfigure}[t]{0.49\textwidth}
        \centering
        \includegraphics[width=\textwidth]{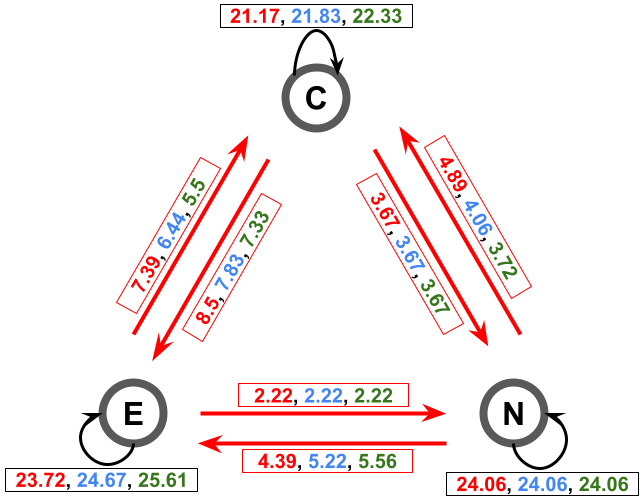}
        \caption{\alphaOne}
    \end{subfigure}
    \hfill
    \begin{subfigure}[t]{0.49\textwidth}
        \centering
        \includegraphics[width=\textwidth]{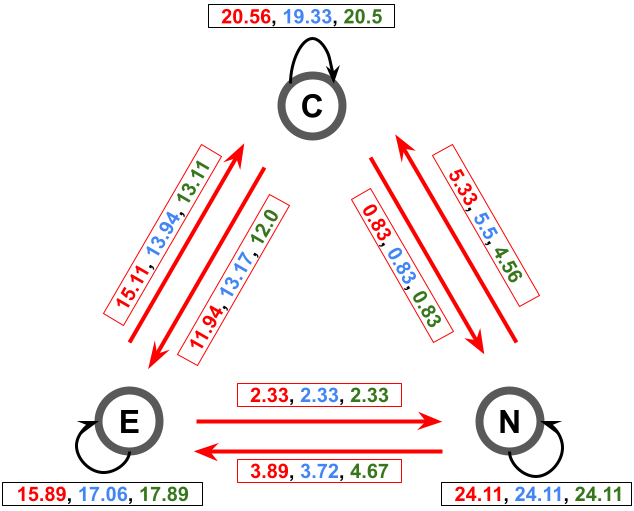}
        \caption{\alphaTwo}
    \end{subfigure}
    \hfill
    \begin{subfigure}[t]{0.49\textwidth}
        \centering
        \includegraphics[width=\textwidth]{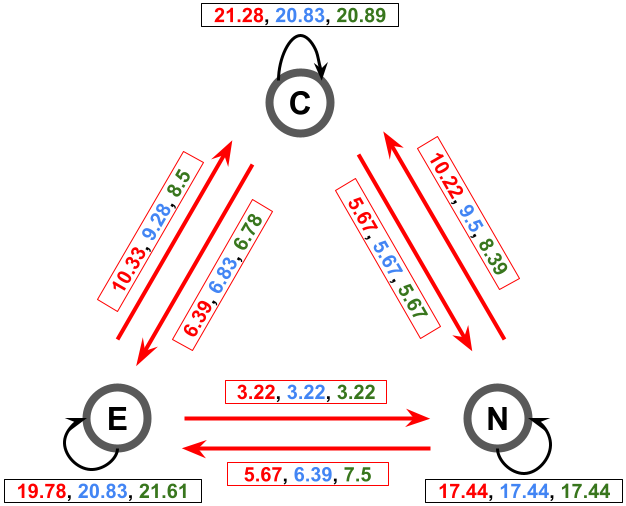}
        \caption{\alphaThree}
    \end{subfigure}
    \caption{Consistency graphs. From left to right the values represent \textcolor{red}{Red} - No Augmentation, \textcolor{mdblue}{Blue} - Orig+Counter, \textcolor{mdgreen}{Green} - MNLI+Orig+Counter.}
    \label{fig:ConfusionMatrix}
\end{figure*}

\begin{table*}[!h]
\small
\centering
\begin{tabular}{lccccccccccc}
\hline
\textbf{Statistics} & \textbf{City}  & \textbf{Album} & \textbf{Person} & \textbf{Movie} & \textbf{Book} & \textbf{F\&D} & \textbf{Org} & \textbf{Paint} & \textbf{Fest} & \textbf{S\&E} & \textbf{Univ} \\
\hline
numerical & 19 & 7 & 28 & 24 & 16 & 8 & 19 & 3 & 14 & 9 & 5 \\
co-reference & 0 & 0 & 2 & 0 & 0 & 0 & 0 & 0 & 0 & 0 & 0 \\
multi-row & 4 & 0 & 15 & 5 & 0 & 7 & 6 & 7 & 1 & 1 & 6 \\
KCS & 21 & 2 & 45 & 9 & 3 & 0 & 24 & 5 & 0 & 5 & 27 \\
temporal & 1 & 6 & 31 & 5 & 1 & 0 & 1 & 4 & 3 & 6 & 2 \\
syntactic-alt & 23 & 0 & 6 & 10 & 2 & 8 & 6 & 2 & 14 & 4 & 28 \\
simple-lookup & 58 & 6 & 54 & 49 & 34 & 19 & 45 & 16 & 32 & 21 & 72 \\
entity-type & 0 & 0 & 54 & 0 & 0 & 0 & 1 & 4 & 0 & 3 & 1 \\
ellipsis & 0 & 0 & 20 & 0 & 0 & 0 & 1 & 0 & 0 & 1 & 0 \\
subjective-oot & 0 & 0 & 0 & 0 & 0 & 0 & 0 & 0 & 0 & 0 & 0 \\
name-identity & 0 & 0 & 6 & 0 & 0 & 0 & 3 & 0 & 0 & 0 & 0 \\
lexical & 25 & 0 & 7 & 20 & 19 & 3 & 30 & 2 & 13 & 6 & 27 \\
quantification & 13 & 5 & 19 & 11 & 11 & 3 & 8 & 1 & 10 & 11 & 3 \\
negation & 0 & 0 & 1 & 0 & 5 & 0 & 5 & 0 & 0 & 0 & 1 \\
\hline
\end{tabular}
\vspace{-0.7em}
\caption{ Distribution of different reasoning types across all categories in \ourdatasetName. }
\label{tab:appendix_reasoning_stat_2}
\vspace{-1.0em}
\end{table*}

\begin{table*}[!h]
\small
\centering
\begin{tabular}{lccccccccccc}
\hline
\textbf{Statistics} & \textbf{City}  & \textbf{Album} & \textbf{Person} & \textbf{Movie} & \textbf{Book} & \textbf{F\&D} & \textbf{Org} & \textbf{Paint} & \textbf{Fest} & \textbf{S\&E} & \textbf{Univ} \\
\hline
\textbf{No. of reasoning} & 164 & 26 & 288 & 133 & 48 & 91 & 149 & 44 & 87 & 67 & 172 \\
\textbf{Avg reasoning} & 2.52 & 1.37 & 2.25 & 1.77 & 1.78 & 1.86 & 1.99 & 2.2 & 1.74 & 1.68 & 2.12 \\
\textbf{Max reasoning} & 4 & 2 & 7 & 3 & 3 & 4 & 4 & 4 & 4 & 3 & 4 \\
\hline
\end{tabular}
\vspace{-0.7em}
\caption{Statistics of reasoning type distribution across the different categories in \ourdatasetName.}
\label{tab:appendix_reasoning_stat_1}
\vspace{-1.0em}
\end{table*}

\begin{table*}[!h]
\small
\centering
\begin{tabular}{lcccc|lcccc}
\hline
\textbf{Reasoning} & \textbf{Average}  & \textbf{Max} & \textbf{Min} & \textbf{Cumulative} & \textbf{Reasoning} & \textbf{Average}  & \textbf{Max} & \textbf{Min} & \textbf{Cumulative}\\
\hline
numerical & 13.82 & 28 & 3 & 152 & entity-type & 5.73 & 54 & 0 & 63 \\
co-reference & 0.18 & 2 & 0 & 2 & ellipsis & 2 & 20 & 0 & 22 \\
multi-row & 4.73 & 15 & 0 & 52 & subjective-oot & 0 & 0 & 0 & 0 \\
KCS & 12.82 & 45 & 0 & 141 & name-identity & 0.82 & 6 & 0 & 9 \\
temporal & 5.45 & 31 & 0 & 60 & lexical & 13.82 & 30 & 0 & 152 \\
syntactic-alt & 9.36 & 28 & 0 & 103 & quantification & 8.64 & 19 & 1 & 95 \\
simple-lookup & 36.91 & 72 & 6 & 406 & negation & 1.091 & 5 & 0 & 12 \\
\hline
\end{tabular}
\vspace{-0.7em}
\caption{Statistics of distribution of different reasoning types across all categories in \ourdatasetName.}
\label{tab:appendix_reasoning_stat_3}
\vspace{-1.0em}
\end{table*}


\textbf{Analysis:} As we observe in Table \ref{tab:appendix_reasoning_stat_2} the cumulative frequency of reasoning types across  each category is highest for Person followed by University and City and the average frequency of reasoning types across category is City followed by Person and Paint. In Table \ref{tab:appendix_reasoning_stat_3} we see that the cumulative frequency of reasoning types across all categories is highest for simple lookup followed by lexical and numerical which have the same frequency.

\section{Consistency Graphs}
\label{sec:appendix-consistency_graphs}
We perform a consistency analysis on three setting, namely No Augmentation, Orig + Count and MNLI + Orig + Count to obtain a better estimate of where pre-training with \ourdatasetName helps improve performance in \datasetName. In Figure \ref{fig:ConfusionMatrix} we have shown the consistency graphs on the 3 settings.

\textbf{Analysis:} We observe in Figure \ref{fig:ConfusionMatrix} that the model is more prone to classifying \contradict as \entail than the other way around in \alphaOne set and there is a significant improvement after pretraining with \ourdatasetName.For \alphaTwo and \alphaThree sets we can see a considerable improvement in \entail being classified as \contradict from pretraining on \ourdatasetName. Pretraining on \ourdatasetName always results in improvements overall.

\end{document}